%% file: main.tex
\documentclass{article} 
\let\paperaddcontentsline\addcontentsline
\usepackage{iclr2027_conference,times}
\let\addcontentsline\paperaddcontentsline

\input{math_commands.tex}

\input{custom_macro.tex}

\usepackage{xurl}

\title{Shockingly Simple Self-retrospection Improves Agentic Models Without RL}

\author{%
\begingroup
\raggedright
\setlength{\parskip}{0pt}
{\normalsize\bfseries
\mbox{Jonathan Light\textsuperscript{1,5\,\Letter}}\enskip
\mbox{Christopher Zhang Cui\textsuperscript{2,5}}\enskip
\mbox{Jeonghye Kim\textsuperscript{3,5}}\enskip
\mbox{Roger Creus Castanyer\textsuperscript{4,5}}\enskip
\mbox{Emiliano Penaloza\textsuperscript{4,5}}\enskip
\mbox{Zhengyan Shi\textsuperscript{5}}\enskip
\mbox{Alessandro Sordoni\textsuperscript{5}}\enskip
\mbox{Marc-Alexandre C\^{o}t\'{e}\textsuperscript{5}}\enskip
\mbox{Xingdi Yuan\textsuperscript{5}}\enskip
\mbox{Minseon Kim\textsuperscript{5}}\endgraf}
\vspace{0.5em}
\mbox{\textsuperscript{1}RPI}\quad
\mbox{\textsuperscript{2}UC San Diego}\quad
\mbox{\textsuperscript{3}KAIST}\quad
\mbox{\textsuperscript{4}Mila}\quad
\mbox{\textsuperscript{5}Microsoft Research}\endgraf
\vspace{0.75em}
\Letter{} Corresponding author:
\href{mailto:jonathan.li.connect@gmail.com}{\nolinkurl{jonathan.li.connect@gmail.com}}\endgraf
\endgroup
}
\date{September 2026}

\arxivpreprint 

\begin{document}
\etocdepthtag.toc{main}
\maketitle

\ifpaperarxiv
\begin{figure}[H]
    \centering
    \includegraphics[width=\linewidth,trim=0 40 0 72,clip]{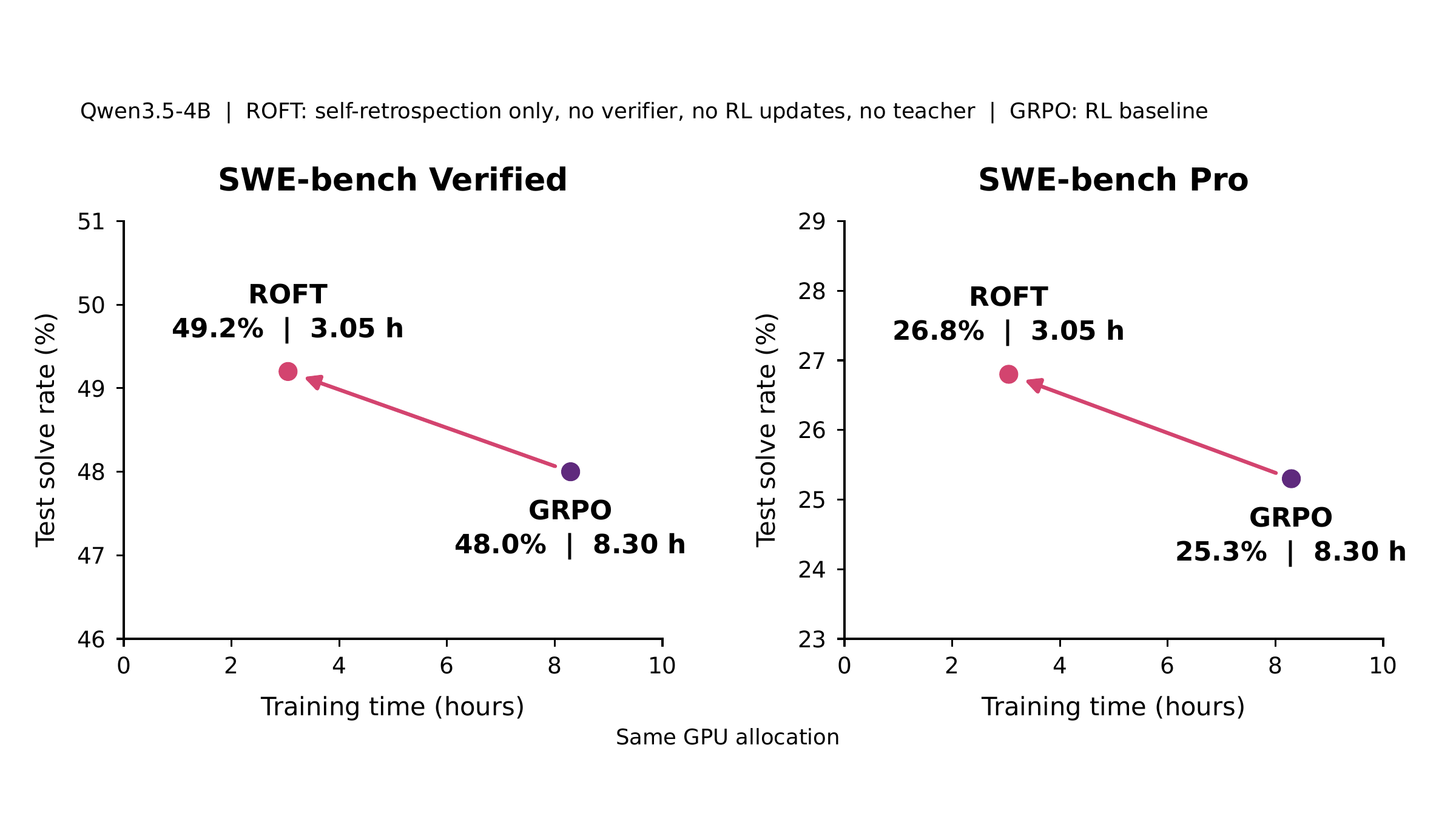}
    \captionsetup{font=small,skip=4pt}
    \caption{\textbf{\methodabbr{} achieves competitive performance with 63\% less training time than GRPO}. Held-out performance of \methodabbr{}
    versus GRPO on agentic coding tasks.}
    \label{fig:retrospection-training-time}
\end{figure}
\fi

\begin{abstract}

\input{sections/abstract}
\end{abstract}

\ifpaperarxiv
\clearpage
\fi

\input{sections/introduction}
\input{sections/background}
\input{sections/methodology}
\input{sections/results}
\input{sections/related_works}
\input{sections/conclusion}
\input{sections/statements}

\bibliography{ref}
\bibliographystyle{iclr2027_conference}

\clearpage
\appendix
\crefalias{section}{appendix}
\etocdepthtag.toc{appendix}
\begingroup
\etocsettagdepth{main}{none}
\etocsettagdepth{appendix}{subsection}
\etocsettocstyle{\section*{Appendix Contents}}{}
\tableofcontents
\endgroup
\clearpage
\input{sections/appendix/related_works_extended}
\input{sections/appendix/experimental_setup}
\input{sections/appendix/baselines}
\input{overleaf_rr_public/retrospection_examples}
\input{sections/appendix/generalization_efficiency}
\input{sections/appendix/beyond_frontier_learning}
\input{sections/appendix/behavioral_induction}
\input{sections/appendix/ablations}

\end{document}

%% file: math_commands.tex
\usepackage{amsmath,amsfonts,bm}

\def\1{\bm{1}}

\DeclareMathAlphabet{\mathsfit}{\encodingdefault}{\sfdefault}{m}{sl}
\SetMathAlphabet{\mathsfit}{bold}{\encodingdefault}{\sfdefault}{bx}{n}



%% file: custom_macro.tex
\usepackage{amsmath}
\usepackage{amssymb}
\usepackage{bbm}
\usepackage{algorithm}
\usepackage{algpseudocode}
\usepackage{graphicx}
\usepackage{subcaption}
\usepackage{wrapfig}
\usepackage{booktabs}
\usepackage{multirow}
\usepackage{tabularx}
\usepackage{microtype}
\usepackage{xcolor}
\usepackage{tikz}
\usetikzlibrary{arrows.meta}
\usepackage{tcolorbox}
\tcbuselibrary{breakable,skins}
\usepackage[utf8]{inputenc}
\usepackage{fvextra}
\DeclareUnicodeCharacter{2192}{\ensuremath{\rightarrow}}
\DeclareUnicodeCharacter{2260}{\ensuremath{\ne}}
\DeclareUnicodeCharacter{2713}{\ensuremath{\checkmark}}
\DeclareUnicodeCharacter{2717}{\ensuremath{\times}}
\usepackage{xspace}
\usepackage{enumitem}
\usepackage{float}
\usepackage{marvosym}
\usepackage{url}
\usepackage{etoc}
\usepackage{hyperref}
\usepackage[nameinlink,capitalise]{cleveref} 

\definecolor{paperPrimary}{HTML}{D3446F}
\definecolor{paperLightOrange}{HTML}{F7976B}
\definecolor{paperLightYellow}{HTML}{FDDC9F}
\definecolor{paperDark}{HTML}{5F2A7D}
\hypersetup{allbordercolors=paperPrimary}
\tcbset{
    paperbox/.style={
        colback=white,coltext=black,colframe=paperPrimary,
        colbacktitle=paperDark,coltitle=white
    },
    retrospection-system/.style={
        colframe=paperDark,colbacktitle=paperDark,coltitle=white
    },
    retrospection-user/.style={
        colbacktitle=paperLightYellow,coltitle=black
    },
    retrospection-target/.style={
        colbacktitle=paperLightOrange,coltitle=black
    }
}

\newif\ifpaperarxiv
\makeatletter
\newcommand{\arxivpreprint}{%
    \paperarxivtrue
    \iclrfinalcopy
    \renewcommand{\@maketitle}{%
        \vbox{\hsize\textwidth
            \lhead{arXiv preprint}%
            {\raggedright\LARGE\scshape\@title\par}%
            \vskip 1.25em
            {\normalfont\normalsize\@author\par}%
            \vskip 1.5em
        }%
    }%
}
\makeatother

\crefname{figure}{Fig.}{Figs.}
\Crefname{figure}{Fig.}{Figs.}
\crefname{table}{Tab.}{Tabs.}
\Crefname{table}{Tab.}{Tabs.}
\crefname{section}{Sec.}{Secs.}
\Crefname{section}{Sec.}{Secs.}
\crefname{appendix}{App.}{Apps.}
\Crefname{appendix}{App.}{Apps.}
\crefname{equation}{Eq.}{Eqs.}
\Crefname{equation}{Eq.}{Eqs.}

\newif\ifcomments
\commentstrue

\newsavebox{\retrospectionexamplebox}
\newenvironment{retrospectionrecord}[2]{%
    \begin{tcolorbox}[enhanced,breakable,paperbox,#2,
        title={#1},title after break={#1\ (continued)},
        fonttitle=\small\bfseries,boxrule=0.5pt,
        left=1.5mm,right=1.5mm,top=1mm,bottom=1mm]
}{\end{tcolorbox}}
\newcommand{\retrospectioninput}[2][]{%
    \VerbatimInput[fontsize=\footnotesize,breaklines,breakanywhere,
        breaksymbolleft={},breaksymbolright={},tabsize=4,#1]{#2}}

\newcommand{\paradigm}{Retrospection Reinforcement\xspace}
\newcommand{\paradigmabbr}{RR\xspace}
\newcommand{\method}{Retrospection-Only Fine-Tuning\xspace}
\newcommand{\methodabbr}{ROFT\xspace}


%% file: sections/abstract.tex
People learn not only by repeating successful actions, but also by
recounting and explaining their experiences, revising their understanding
to guide future behavior. Can a language-model agent improve its future actions by training
only on explanations of its own experience? We investigate this question
by studying \emph{\method{}} (\methodabbr{}), a minimal online
procedure designed to isolate the effect of explanation-only training
on subsequent behavior. The agent attempts a task, observes available feedback,
generates a retrospective explanation, and is fine-tuned with a
next-token prediction loss on the \textit{explanation tokens alone}. The
procedure uses neither an external teacher nor a reward-based policy
update. In software-engineering experiments with Qwen3.5-4B,
\methodabbr{} is trained on problems with mixed successful and
unsuccessful base-model attempts. On held-out SWE-bench Verified and Pro,
it reaches 49.2\% and 26.8\% solve rates after 20 updates without using a verifier, compared with
GRPO's 48.0\% and 25.3\% after 40 updates in the evaluated runs, and
makes faster early progress in training time and sampled attempts.
It also learns to solve individual tasks on which all 64 sampled
base-model attempts failed, showing that learning can begin without any
initially successful trajectories.
Behavioral analyses find that \methodabbr{} indirectly assigns credit to actions, encouraging good actions and discouraging incorrect ones.
Moreover, prompting retrospections to emphasize more
direct solutions yields shorter subsequent attempts even without an
explicit length penalty. Together, these findings show that learning to
explain can also improve learning to do, establishing self-generated
retrospections as useful training targets 
and
motivating further study of explanation-to-action transfer.

%% file: sections/introduction.tex
\section{Introduction}
\label{sec:introduction}


We sometimes understand an experience differently in the act of recounting
it. Explaining why a conversation went badly, someone might begin with
``I did not make my point clearly enough.'' But as they reconstruct the
exchange, another explanation emerges: they kept defending their proposal
while the other person was questioning its premise. The lesson is no
longer to explain the same point more forcefully, but to establish which
question needs answering. Nothing about the original outcome has changed.
What changes is their understanding of what happened---and, with it, how
they might act next time. For language-model agents, this suggests a
complementary training target: not only the actions taken during an
attempt, but retrospective explanations that make sense of the experience.

Reinforcement learning with verifiable rewards (RLVR) offers an effective
way to learn from experience by reinforcing task-solving behavior
according to its outcomes \citep{lambert2024tulu3,deepseekai2025deepseekr1}.
Methods such as GRPO use relative rewards
among sampled attempts to update the policy \citep{shao2024deepseekmath}.
Yet a single outcome reward, success or failure, does not identify which
assumption was mistaken, which decision mattered, or when a correction
should apply. This limitation is especially relevant when rewards are
sparse: an all-failure group has no within-group binary-reward contrast,
even though its trajectories may contain useful evidence. 

Natural
language can express an interpretation of that evidence, connecting
observations to decisions and stating lessons that may apply elsewhere.
Prior work has used reflections in several ways: as context for later attempts
\citep{shinn2023reflexion}, or to obtain successful reasoning and
reflection-informed solutions for training \citep{zelikman2022star,shi2026erl}.
Critique fine-tuning, which trains on critiques as a form of reflection, demonstrates that teacher critiques can be transferred offline to a student to improve its question-answering performance \citep{wang2025cft,wang2026scft}.
However, it remains unclear whether training only on self-generated reflection can improve agents. 
The central question is therefore:

\begin{tcolorbox}[
    paperbox,
    boxrule=0.7pt,
    left=3mm,right=3mm,top=2mm,bottom=2mm,
    before skip=8pt,after skip=8pt
]
    \centering\bfseries
    Can an agent improve its future actions by training only on
    self-generated retrospective explanations of its own experience?
\end{tcolorbox}

We call this direction \textbf{\paradigm{} (\paradigmabbr{})}:
training on self-generated retrospections of past experience, with the aim
of improving subsequent behavior. A retrospection may summarize events,
track how beliefs evolved, explain decisions or outcomes, identify
corrections, or articulate lessons for future attempts. Our procedure trains on this retrospective
text rather than directly supervising the recorded task-solving actions.
The hypothesis is that
learning to explain an experience can improve how the model acts in later attempts. We assess this
hypothesis through subsequent task performance and behavior changes, not through
the fluency of explanations or a reduction in their prediction loss.

To study this route in isolation, we introduce
\textbf{\method{} (\methodabbr{})}, a deliberately minimal online
procedure. The agent attempts a task, observes the outcome and available
feedback, and generates a retrospection of its own attempt. We then
fine-tune the same model with a next-token prediction loss on the
retrospection tokens only. The task, attempted actions, and observations
serve as context, not prediction targets. Both successful and unsuccessful
attempts are eligible, and the procedure uses neither an external teacher
nor a reward-based policy update. Subsequent attempts receive no stored
retrospection and require no additional reflection step: any benefit must
transfer through the updated weights. These exclusions make simplicity
an experimental choice, isolating what retrospection training can
contribute without direct action supervision.

Our software-engineering experiments provide evidence that learning to
explain can improve learning to do. First, when trained on problems
with both successful and unsuccessful base-model attempts---the
\emph{learning zone}---\methodabbr{} achieves higher held-out solve rates
than GRPO at the reported checkpoints, with faster early progress in
wall-clock time and sampled solution attempts
(\cref{sec:training_efficiency}), without a verifier. Second, it learns to solve individual
tasks on which all 64 sampled base-model attempts failed, extending
learning \emph{beyond the frontier} without an initially
successful trajectory or variation in binary outcomes
(\cref{sec:empirical_frontier}). Third,
behavioral analyses find changes consistent with indirect credit
assignment: likelihood increases are more selective for correct than
incorrect recorded turns. Changing the retrospection prompt to emphasize
more direct solutions also produces shorter subsequent attempts, despite
the absence of an explicit length penalty or action-target loss
(\cref{sec:behavorial_analysis}).

These results suggest a complementary design axis for agent training:
not only which experiences to learn from, but what to learn to say about
them. Our results show that experience can support useful training
targets even when binary outcomes provide no contrast or in the absence
of any outcome verdict. The prompt-dependent behavioral changes further
suggest that the content of those targets matters: changing what an
agent emphasizes in retrospection can change how it subsequently acts.
Retrospection is therefore not merely a record of experience, but a
potentially steerable source of supervision.

%% file: sections/background.tex



%% file: sections/methodology.tex
\section{Retrospection-Only Training}
\label{sec:methodology}

An unsuccessful attempt may be a poor example to imitate but still
provide useful experience to interpret. Here, we describe how an agent can
be trained exclusively on retrospections of its own experience and
evaluated on its subsequent task performance.

\subsection{Learning to explain, evaluated by doing}

\emph{Learning to do} directly optimizes task-solving outputs, including
reasoning, answers, and actions. Reward-based updates, imitation, and
distillation can all serve this purpose. \emph{Learning to explain}
instead trains the model to produce a retrospective account of a completed attempt.
We call this direction \textbf{\paradigm{} (\paradigmabbr{})}.

A \emph{retrospection} is a textual account of past experience, generated
in light of the recorded interaction and any available feedback.
Its form can range from a trajectory summary or a ledger of evolving
beliefs to an explanation of decisions or outcomes, a correction, or a
reusable lesson. 
No particular content structure is required.
Retrospections may discuss actions and possible corrections; the
distinction is between training on a post-attempt account and directly
supervising task-solving actions.
\Cref{fig:learning-signals} contrasts direct action training with
retrospection prediction, where the recorded attempt supplies context
rather than prediction targets. Our hypothesis is that learning to explain an experience changes the
knowledge and representations available when the same model next acts.
Below, we describe a minimal algorithm for studying this hypothesis.

\input{figures/methodology}

\vspace{-0.1in}
\subsection{\methodabbr{}: A minimal test of explanation-to-action transfer}
\label{sec:retrospection}

\textbf{\method{} (\methodabbr{})} repeats a simple loop: attempt a
task, explain the experience, fit the explanation, and act again using
the updated weights. We exclude direct action supervision and retained
retrospection so that neither can account for an improvement in doing.

\vspace{-0.05in}
\paragraph{Collect experience.}
The agent attempts a task $x$, producing an interaction trajectory
$\tau$, and receives available feedback $v$. In our software-engineering
experiments, the record includes actions and observations, the submitted
patch, and available test feedback. Both successful and unsuccessful
attempts are eligible: the method does not require a successful example
to imitate.

\vspace{-0.05in}
\paragraph{Explain the experience.}
The same model samples $K$ retrospections from a context $h(x,\tau,v)$
containing an instruction and a bounded rendering of the record.
Our default experimental prompt asks for a consequential assumption or
decision, supporting or contradicting evidence, a correction when
appropriate, and concrete triggers for applying the lesson. This is one instantiation
of retrospection, not a requirement on its form. The model supplies its
own interpretations rather than receiving an external teacher's
critiques. Available verdicts condition these interpretations; they do
not become reward weights in the objective. 
The example below shows how even a passing attempt can yield a lesson about a specific decision.

\begin{lrbox}{\retrospectionexamplebox}
\begin{minipage}{\linewidth}
\begin{tcolorbox}[
    paperbox,
    colframe=black,
    colbacktitle=black,
    title={Retrospection after a passing attempt (full prompts and responses in \cref{app:retrospection_examples})},
    fonttitle=\small\bfseries,
    fontupper=\small,
    left=2mm,right=2mm,top=1mm,bottom=1mm
]
\colorbox{paperDark}{\textcolor{white}{\textbf{System prompt (abridged).}}}\par
Identify a consequential
assumption or decision, the evidence for or against it, a correction if
needed, and concrete triggers for applying the lesson.
State uncertainty when evidence is insufficient.

\smallskip
\colorbox{paperLightYellow}{\textcolor{black}{\textbf{User prompt (abridged).}}}\par
\textbf{Task:} Prevent a failed Pact context-manager test from leaving
stale interactions that cause later tests to fail.\par
\textbf{Action:} Inspect the context manager's \texttt{\_\_exit\_\_}
and \texttt{verify()} methods.\par
\textbf{Observation:} \texttt{verify()} clears the interaction list,
but \texttt{\_\_exit\_\_} skips it when an exception occurs.\par
\textbf{Action:} Clear \texttt{self.\_interactions} on exceptional exit,
then run the Pact and consumer tests.\par
\textbf{Observation:} All 70 tests pass.

\smallskip
\colorbox{paperLightOrange}{\textcolor{black}{\textbf{Retrospection (final-answer excerpt).}}}\par
``The correct decision was to clear interactions
directly in \texttt{\_\_exit\_\_} when an exception is detected, ensuring
proper state reset regardless of whether \texttt{verify()} is called.''

\end{tcolorbox}
\end{minipage}
\end{lrbox}
\ifdim\dimexpr\ht\retrospectionexamplebox+\dp\retrospectionexamplebox\relax>0.5\textheight
    \PackageError{retrospection-example}{Main-text example exceeds half a page}{Shorten the example.}
\fi
\typeout{RETROSPECTION-BOX-HEIGHT: \the\dimexpr\ht\retrospectionexamplebox+\dp\retrospectionexamplebox\relax;
    HALF-PAGE-LIMIT: \the\dimexpr\textheight/2\relax}
\noindent\usebox{\retrospectionexamplebox}

\paragraph{Fit only the explanation, not the attempt.}
Let $\mathcal{D}_t$ contain the retained context--retrospection pairs
$(h,y)$ for update $t$. Holding these generated targets fixed, we
fine-tune the model $\pi_\theta$ with next-token cross-entropy:
\vspace{-0.1in}
\begin{equation}
    \mathcal{L}_{\mathrm{\methodabbr{}}}(\theta;\mathcal{D}_t)
    =
    -\frac{1}{T_t}
    \sum_{(h,y)\in\mathcal{D}_t}
    \sum_{j=1}^{|y|}
    \log \pi_\theta(y_j \mid h,y_{<j}),
    \qquad
    T_t=\sum_{(h,y)\in\mathcal{D}_t}|y|.
    \label{eq:ro}
\end{equation}
The loss averages globally over retained retrospection tokens.
The task,
attempted actions, observations, and feedback are masked as prediction
targets, although gradients can flow through their context
representations. There is no action-target loss, reward-based policy
update, or independent semantic-quality filter on the retrospections.

\vspace{-0.1in}
\paragraph{Act again using updated weights.}
The learner collects fresh attempts and retrospections as training
proceeds. Subsequent attempts, including evaluation, receive neither
stored retrospection text nor an added retrospection step.
Any benefit must therefore transfer through the updated weights,
rather than through access to a written lesson. This closes the loop
between explaining and doing while keeping their training targets
distinct. Sampling settings, target processing, batch construction,
and asynchronous weight refresh are specified in
\cref{app:experimental_setup}.

%% file: figures/methodology.tex
\input{overleaf_rr_public/assets/rr_ar_concepts/figure_spotlight}

%% file: overleaf_rr_public/assets/rr_ar_concepts/figure_spotlight.tex
\newsavebox{\methodologyfigurebox}
\newlength{\methodologyfigureheight}
\begin{figure}[ht]
    \begin{lrbox}{\methodologyfigurebox}
    \begin{minipage}{\linewidth}
    \centering
    \includegraphics[width=\linewidth]{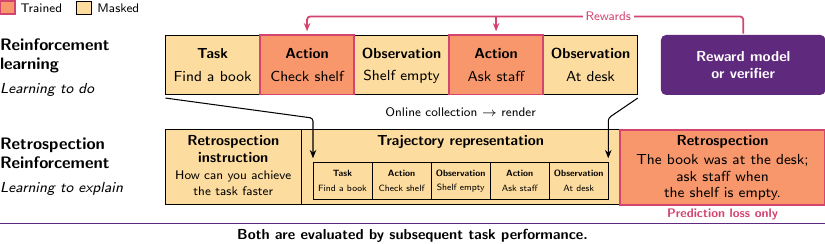}
    \captionsetup{font=footnotesize,skip=3pt}
    \caption{Illustrative RL versus RR. Black arrows show RR's online collection
    and rendering, not RL updates.
    Orange marks targets; yellow, masked context.}
    \label{fig:learning-signals}
    \label{fig:rr-ar-spotlight}
    \end{minipage}
    \end{lrbox}
    \setlength{\methodologyfigureheight}{\dimexpr
        \ht\methodologyfigurebox+\dp\methodologyfigurebox+\textfloatsep\relax}
    \typeout{METHODOLOGY-FIGURE-WIDTH: \the\wd\methodologyfigurebox;
        HEIGHT-WITH-SPACING: \the\methodologyfigureheight;
        QUARTER-PAGE-LIMIT: \the\dimexpr\textheight/4\relax}
    \ifdim\methodologyfigureheight>0.25\textheight
        \PackageError{methodology-figure}{Figure and caption exceed quarter-page budget}
            {Shorten the diagram or caption, including the reserved float separation.}
    \fi
    \centering\usebox{\methodologyfigurebox}
\end{figure}

%% file: sections/results.tex
\vspace{-0.05in}
\section{Results}
\label{sec:preliminary}
\label{sec:results}
\vspace{-0.05in}

We evaluate explanation-to-action transfer through three questions.
Does retrospection-only training improve held-out solve rates, and at
what training cost?
Can learning begin when all sampled base model attempts fail?
Does retrospection content shape subsequent behavior as measured by action-likelihood changes, interventions on retrospection
instructions, and changes to content weighting?
These questions assess behavioral transfer. 

\begin{figure}[htbp]
    \centering
    \includegraphics[width=0.9\textwidth]{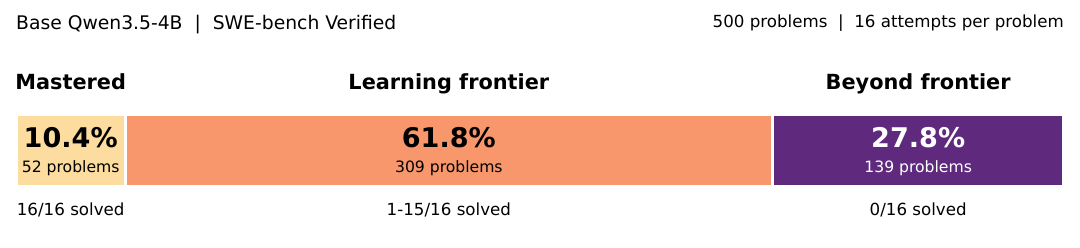}
    \caption{\textbf{SWE-bench Verified problems categorized by base model outcomes over 16 attempts.}
    Percentages of problems solved in all, some, or none of 16 attempts
    pooled from 16 evaluations.}
    \label{fig:frontier-zones}
\end{figure}

We distinguish two learning regimes using sampled base-model outcomes.
A problem is in the \textbf{learning zone} if the base model produces
both successful and unsuccessful solutions across $k$ sampled attempts;
it is \textbf{beyond frontier} if none of those attempts
succeeds. 
\Cref{fig:frontier-zones} illustrates the distinction for Qwen3.5-4B
on all 500 SWE-bench Verified problems using 16 attempts per problem.
The experiments below use three attempts to screen the learning-zone
training subset and 64 attempts to establish the all-failure starting
points for single-task training.
\input{overleaf_rr_public/assets/frontier_zones/metrics}

\vspace{-0.1in}

\vspace{-0.05in}
\subsection{Learning Zone}
\vspace{-0.05in}
\label{sec:training_efficiency}
\input{overleaf_rr_public/assets/training_efficiency/metrics}

We first evaluate generalization and efficiency in the learning zone by comparing
\methodabbr{} and GRPO initialized from Qwen3.5-4B and trained on
SWE-rebench-767, a subset of 767 problems from SWE-rebench
\citep{badertdinov2025swerebench}. 
Our training subset is \emph{specifically
selected to favor GRPO}: because its reward-derived policy-gradient
signal \emph{requires} variation in rewards within a group, we retain only
problems on which Qwen3.5-4B produces both successful and unsuccessful
solutions across three initial attempts.
We assess held-out performance on SWE-bench
Verified \citep{openai2024swebenchverified} and SWE-bench Pro
\citep{deng2025swebenchpro}. All three datasets require agents to resolve
real-world repository issues by producing code patches evaluated against
executable tests. SWE-bench Verified comprises 500 human-validated
Python tasks, whereas SWE-bench Pro emphasizes more complex,
long-horizon tasks that often require substantial changes across
multiple files.  Qwen3.5-4B performance is 44.2\% and 23.9\% on Verified and Pro respectively in our setup. 

\begin{figure}[!t]
    \centering
    \vspace{-0.2in}
    \begin{subfigure}[t]{0.65\textwidth}
        \includegraphics[width=\linewidth]{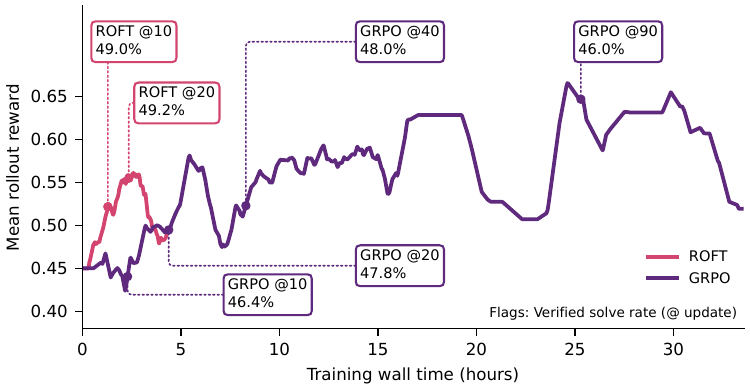}
        \caption{Training reward and held-out evaluations.}
        \label{fig:extended-training-checkpoints}
    \end{subfigure}\hfill%
    \begin{subfigure}[t]{0.325\textwidth}
        \includegraphics[width=\linewidth]{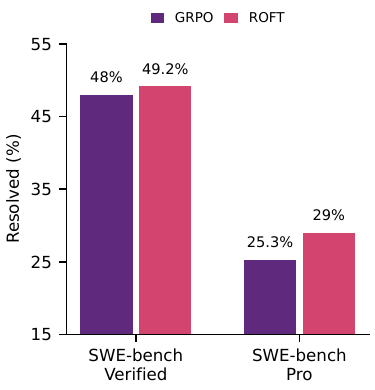}
        \caption{Held-out performance.}
        \label{fig:checkpoint-benchmark-bars}
    \end{subfigure}
    \caption{\textbf{Early training gains transfer to held-out tasks, while later training reward need not improve generalization.}
    (a) Training reward through 40 \methodabbr{} updates and 100 GRPO
    updates. Flags report separate SWE-bench
    Verified test set solve rates at the marked checkpoints.
    (b) Solve rates on SWE-bench Verified and SWE-bench Pro for
    \methodabbr{} at update 20 and GRPO at update 40.}
    \vspace{-0.1in}
    \label{fig:checkpoint-comparison}
\end{figure}

\vspace{-0.05in}
\paragraph{Generalization.}
\methodabbr{} achieves higher held-out solve rates on both
benchmarks in \cref{fig:checkpoint-benchmark-bars}: its
20-update checkpoint reaches 49.2\% on SWE-bench Verified and 29.0\%
on SWE-bench Pro, compared with 48.0\% and 25.3\% for GRPO at
40 updates. These gains do not require continued improvement in
training reward. \methodabbr{}'s training curve levels off around
20 updates, whereas GRPO's reward continues
to improve with longer training. Yet GRPO's Verified solve rate falls
from 48.0\% at update 40 to 46.0\% at update 90
(\cref{fig:extended-training-checkpoints}), a pattern consistent
with overfitting to the training tasks.
We hypothesize that \methodabbr{}'s early saturation reflects
increasingly repetitive retrospections as training revisits familiar
tasks, reducing the new supervision they provide, which is consistent with decreasing reflection entropy and SFT loss. See \Cref{app:generalization_efficiency} for details.

\begin{figure}[!t]
    \centering
    \begin{subfigure}[t]{0.32\textwidth}
        \includegraphics[width=\linewidth]{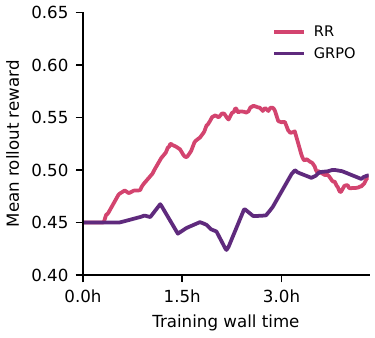}
        \caption{Training wall time.}
        \label{fig:efficiency-time}
    \end{subfigure}\hfill%
    \begin{subfigure}[t]{0.32\textwidth}
        \includegraphics[width=\linewidth]{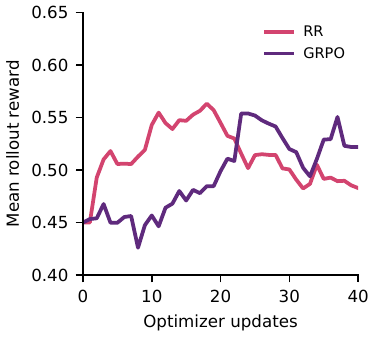}
        \caption{Optimizer updates.}
        \label{fig:efficiency-updates}
    \end{subfigure}\hfill%
    \begin{subfigure}[t]{0.32\textwidth}
        \includegraphics[width=\linewidth]{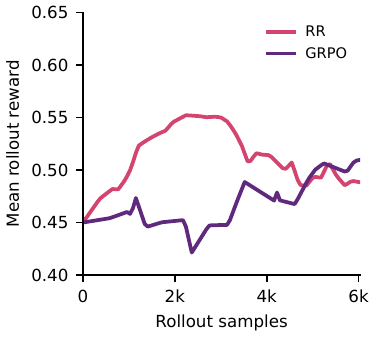}
        \caption{Sampled solution attempts.}
        \label{fig:efficiency-samples}
    \end{subfigure}
    \caption{\textbf{\methodabbr{} has higher training efficiency.}
    \methodabbr{} and GRPO trained on SWE-rebench, with training
    reward plotted against (a) wall time, (b) optimizer updates, and
    (c) completed solution attempts, including zero-advantage attempts rejected by
    GRPO. 
    See \cref{app:generalization_efficiency} for details.
    }
    \vspace{-0.1in}
    \label{fig:training-efficiency}
\end{figure}

\vspace{-0.1in}
\paragraph{Training efficiency.}
Starting from Qwen3.5-4B and training on SWE-rebench-767,
\methodabbr{} makes faster early progress than GRPO in terms of
wall-clock time, optimizer updates, and sampled solution attempts
(\cref{fig:training-efficiency}).
With the same allocation of four training and four inference GPUs,
\methodabbr{} completes 40 updates in \EfficiencyRRHours{} hours
using \EfficiencyRRSamples{} solution attempts, compared with
\EfficiencyGRPOHours{} hours and \EfficiencyGRPOSamples{} attempts
for GRPO, i.e. approximately half the training time and half as many
solution attempts.
This early advantage also transfers to held-out performance:
\methodabbr{} solves 49.0\% of SWE-bench Verified after ten updates
and 1.29 hours, exceeding the 48.0\% achieved by GRPO's
40-update checkpoint, which takes 8.30 hours
(\cref{fig:extended-training-checkpoints}).
Thus, \methodabbr{} attains a higher observed test score
in roughly one-sixth the training time.

\vspace{-0.05in}
\paragraph{Comparison with prior methods.}
We also compare \methodabbr{} with baselines from prior work, with results
in \cref{fig:prior-work-baselines} and method details in \cref{sec:baselines}.
\input{overleaf_rr_public/assets/combined_performance/metrics}
\methodabbr{} can also be effectively combined with GRPO by adding a retrospection step to each GRPO rollout (\cref{fig:combined-performance}).


\vspace{-0.05in}
\subsection{Learning Beyond the Model's Frontier}
\vspace{-0.05in}
\label{sec:empirical_frontier}
\input{overleaf_rr_public/assets/hard_tasks/metrics}

We next ask whether learning can begin without any observed successful
base-model attempts, rather than only on the mixed-success problems. With binary success rewards, an all-failure GRPO group has
zero within-group advantage and provides no reward-derived
policy-gradient signal, so GRPO does not learn in this regime. \methodabbr{} instead obtains supervision from
retrospections of failed attempts.

\begin{figure}[htbp]
    \centering
    \captionsetup[subfigure]{justification=centering}
    \begin{subfigure}[t]{0.32\textwidth}
        \includegraphics[width=\linewidth]{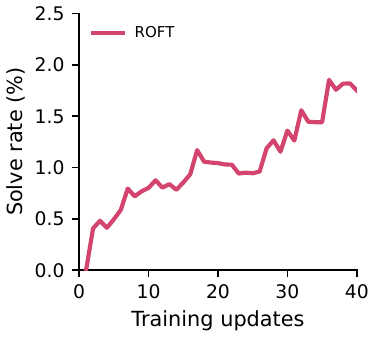}
        \caption{SymPy 20916\\(SWE-bench Verified).}
        \label{fig:zero-advantage-sympy}
    \end{subfigure}\hfill%
    \begin{subfigure}[t]{0.32\textwidth}
        \includegraphics[width=\linewidth]{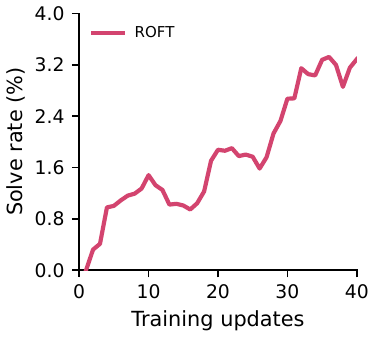}
        \caption{SymbiFlow 17\\(SWE-rebench).}
        \label{fig:zero-advantage-symbiflow}
    \end{subfigure}\hfill%
    \begin{subfigure}[t]{0.32\textwidth}
        \includegraphics[width=\linewidth]{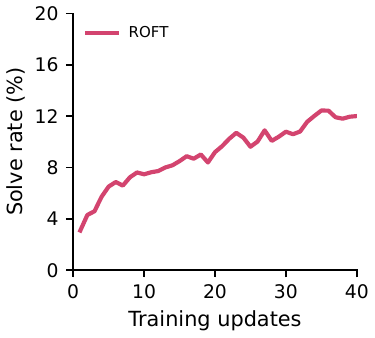}
        \caption{Pylint 6386\\(SWE-bench Verified).}
        \label{fig:low-pass-rate-pylint}
    \end{subfigure}
    \caption{\textbf{\methodabbr{} can improve performance on near and beyond frontier problems.}
    Online solve rates during 40 updates of \methodabbr{} on a single task.
    SymPy and SymbiFlow begin with 0/64 successes; Pylint begins with 2/64
    and reaches a smoothed solve rate of \HardPylintEarlyFinal{} at update 40.
    We see similar results on Django and with longer training runs (\cref{app:additional_task_learning}).}
    \label{fig:zero-advantage-curves}
\end{figure}

To study an all-failure starting point, we train separate Qwen3.5-4B
runs on individual SWE-bench Verified and SWE-rebench tasks, each beginning with
0/64 successful base model attempts on that task and using only self-generated
retrospections for supervision.
After 40 updates, the online solve rate reaches \HardSympyFinal{} on SymPy and
\HardSymbiflowFinal{} on SymbiFlow
(\cref{fig:zero-advantage-curves}). Thus, retrospection-only training can produce successful solutions from
an all-failure starting point. Training on a single problem with
\methodabbr{} does not degrade test-set performance significantly
(\cref{fig:single-task-test-performance}).

\vspace{-0.1in}
\subsection{Behavioral Analysis}
\vspace{-0.05in}
\label{sec:behavorial_analysis}

\input{overleaf_rr_public/assets/step_directions/metrics}
\input{overleaf_rr_public/assets/faster_prompt/metrics}

\begin{figure}[!htb]
    \centering
    \vspace{-0.1in}
    \begin{subfigure}[t]{0.32\textwidth}
        \begin{minipage}[c][45mm][c]{\linewidth}
            \includegraphics[width=\linewidth]{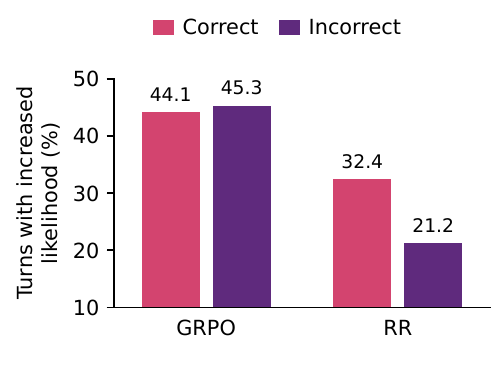}
        \end{minipage}
        \caption{Turn likelihood increases.}
        \label{fig:credit-assignment}
    \end{subfigure}\hfill%
    \begin{subfigure}[t]{0.32\textwidth}
        \begin{minipage}[c][45mm][c]{\linewidth}
            \includegraphics[width=\linewidth]{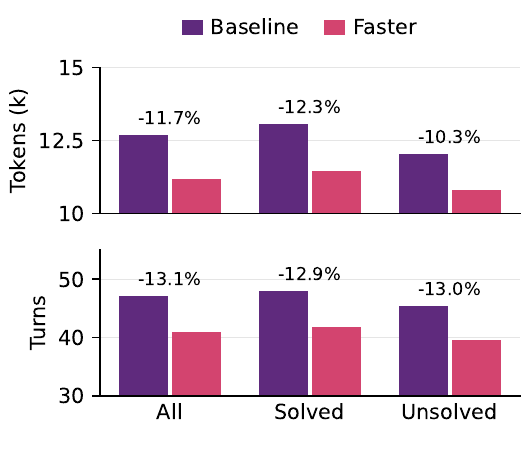}
        \end{minipage}
        \caption{Rollout lengths at update 10.}
        \label{fig:faster-rollouts}
    \end{subfigure}\hfill%
    \begin{subfigure}[t]{0.32\textwidth}
        \begin{minipage}[c][45mm][c]{\linewidth}
            \includegraphics[width=\linewidth]{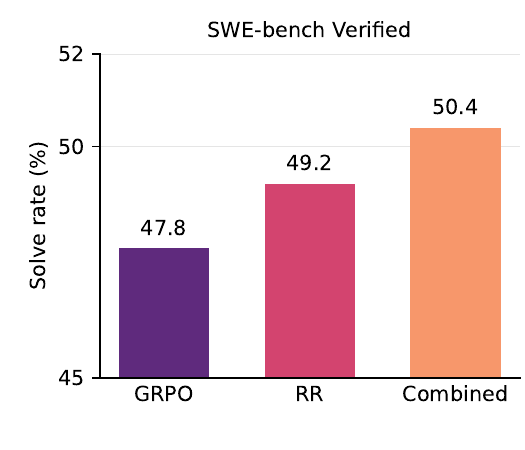}
        \end{minipage}
        \caption{\methodabbr{} combined with GRPO.}
        \label{fig:combined-performance}
    \end{subfigure}
    \caption{
    \textbf{(a) \paradigmabbr{} conducts more credit assignment than GRPO.}
    Bars show trajectory-balanced fractions of recorded turns whose mean
    token log-likelihood increases relative to the initial model after
    ten optimizer updates. Under \paradigmabbr{}, correct turns increase
    in likelihood more often than incorrect turns; GRPO shows little
    separation. This pattern is consistent with indirect credit assignment
    from retrospection-only training.
    We attribute the lower likelihood-increase fractions under
    \paradigmabbr{} to its larger policy shift over the same ten updates.
    \textbf{(b) We can induce behaviors such as completing the task faster through \paradigmabbr{} alone without any length penalties.} Mean tokens and turns per trajectory under the baseline \methodabbr{} prompt and a variant that prompts the model to retrospect on how to solve the task faster. \textbf{(c) \methodabbr{} can be combined with GRPO.} Update 20 performance. 
    See \cref{app:credit_assignment,app:faster_rollouts}.}
    \label{fig:behavioral-analysis}
\end{figure}

\paragraph{Credit assignment.}
\label{sec:credit_assignment}
Can retrospection-only training assign credit to individual actions? We evaluate a fixed set of
base model trajectories on SWE-rebench-767, with assistant turns
labeled for correctness by GPT-6 Astra using the recorded evidence.
For each turn, including both reasoning and actions, we compare its
mean token log-likelihood under the initial Qwen3.5-4B model and after
ten optimizer updates of \methodabbr{} or GRPO, conditioning on the same recorded history.
\Cref{fig:credit-assignment} shows the fraction of turns whose
likelihood increases, computed separately for each correctness label
within each trajectory and then averaged equally across trajectories
containing that label.
Under \paradigmabbr{}, correct turns increase in likelihood more
often than incorrect turns (\StepRRMainCorrectIncreased{} versus
\StepRRMainIncorrectIncreased{}), whereas GRPO shows similar rates
for both (\StepGRPOMainCorrectIncreased{} versus
\StepGRPOMainIncorrectIncreased{}).
This separation is consistent with indirect credit assignment
without direct action supervision.
See \cref{app:credit_assignment} for details.

\vspace{-0.1in}
\paragraph{Inducing behavioral changes through retrospection.}
Can the content of retrospection steer subsequent task-solving behavior?
We compare standard \methodabbr{} with a variant that asks the model to
identify how it could have solved the task more directly after a
successful attempt while preserving correctness.
Both runs use the same initialization, training data, solver prompt,
and retrospection-only objective.
The variant uses
\FasterAllTokenReduction{} fewer generated assistant tokens and
\FasterAllTurnReduction{} fewer assistant turns than the baseline (\cref{fig:faster-rollouts}).
These observations are consistent with content-directed behavioral
change from training on retrospections alone, without an explicit
length penalty or action-target supervision.
\Cref{app:faster_rollouts} gives the experimental procedure,
complete prompts, and detailed results.

\begin{figure}[t]
    \centering
    \vspace{-0.2in}
    \captionsetup[subfigure]{justification=centering}
    \begin{subfigure}[t]{0.32\textwidth}
        \includegraphics[width=\linewidth]{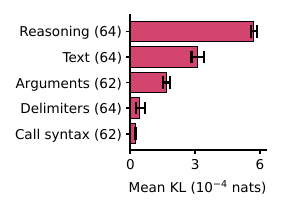}
        \caption{KL conditioned on role.}
        \label{fig:reflection-rollout-roles}
    \end{subfigure}\hfill%
    \begin{subfigure}[t]{0.32\textwidth}
        \includegraphics[width=\linewidth]{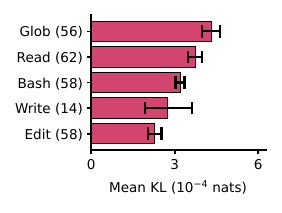}
        \caption{KL conditioned on tool.}
        \label{fig:reflection-tools}
    \end{subfigure}\hfill%
    \begin{subfigure}[t]{0.32\textwidth}
        \includegraphics[width=\linewidth]{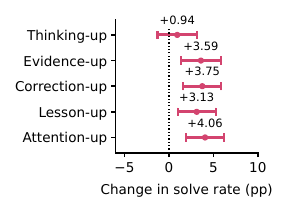}
        \caption{Retrospection impact.}
        \label{fig:reflection-weighting-performance}
    \end{subfigure}
    \caption{
    \textbf{(a, b) Reasoning and read/search contexts show larger prediction
    changes.}
    KL divergence on fixed trajectories before and after one \methodabbr{}
    update, grouped by token roles and tools.
    \textbf{(c) Evidence, correction, and attention weighting yield the largest
    observed solve-rate gains.}
    Solve-rate changes relative to uniform weighting (Uniform) after one update
    with doubled category weights
    or continuous prompt-attention weights (Attention-up), normalized to mean one.
    Points show means with $\pm1$ bootstrap SE; evaluation uses ten repetitions
    on each of the 64 training problems (640 attempts per model).
    Details in \cref{app:reflection_token_analysis}.}
    \vspace{-0.1in}
    \label{fig:reflection-token-analysis}
\end{figure}

\paragraph{Fine-grained analysis of action-trajectory changes.}
Where does learning to explain change the model's subsequent action
predictions? We isolate a single retrospection-only update of Qwen3.5-4B
on 255 retrospections from 64 source attempts, then compare the full
next-token distributions before and after training on the same original
rollout prefixes. Among the mechanically identified roles,
reasoning has the largest task-balanced mean forward KL,
a $3.45\times$ difference from tool calls
(\cref{fig:reflection-token-analysis}).
Changes are also concentrated earlier in the rollout: the first
normalized assistant-turn decile has the largest mean KL
(\cref{fig:reflection-progress}).
Read/search contexts shift more than modification contexts:
the task-balanced mean KL for \texttt{Glob} and \texttt{Read} is
$1.35$--$1.89$ times that for \texttt{Write} and \texttt{Edit}
across the four pairwise comparisons.
These patterns are consistent with transfer to reasoning and information
gathering despite the absence of action-target supervision. 
See \Cref{app:reflection_token_analysis} for details.

\vspace{-0.05in}
\paragraph{Weighting retrospection contents.}
Does emphasizing particular retrospection content improve subsequent task solving?
Using the same frozen retrospections, we compare uniform target weights
(Uniform) with variants that double the relative weight of
evidence, task-specific corrections, reusable lessons, or thinking tokens.
GPT-6 Astra labels the evidence, correction, and lesson categories;
thinking tokens are identified mechanically.
Attention-up instead weights each nonstructural target token by one plus
the base model's mean attention mass to the user prompt when predicting it.
All variants receive one update from the same base model, with weights
normalized to mean one across the target batch.
We evaluate each checkpoint on the same 64 training problems with ten
paired evaluation seeds, yielding 640 attempts per model.
Attention-up, correction-up, and evidence-up have the highest observed
solve rates among these variants: 54.69\%, 54.38\%, and 54.22\%,
respectively, versus 50.63\% for Uniform
(\cref{fig:reflection-token-analysis}).

\paragraph{Why might retrospection training work?}
We hypothesize that \paradigmabbr{} improves behavior by refining the
model's understanding of an experience rather than directly reinforcing
individual actions. Revising an assumption or learning when a correction
applies may influence many future decisions, providing a route to
generalization and potentially explaining the faster early progress
relative to GRPO. Such higher-level supervision can also be more
fine-grained than a trajectory-level reward: evidence grounds an
explanation in observed behavior, while corrections identify which
decisions should change and why. Correct and incorrect turns show a
larger gap in likelihood-increase rates under \paradigmabbr{} than under
GRPO, consistent with more selective credit assignment
(\cref{fig:credit-assignment}). Mechanistically, predicting
retrospection tokens can backpropagate gradients through attention to
representations of the recorded actions and observations, updating
shared model parameters even though the trajectory tokens themselves
carry no prediction loss. This provides a pathway for explanation
training to change subsequent actions without directly supervising them.
Consistent with this, giving greater weight to retrospection
tokens whose predictions attend more strongly to the user prompt
containing the trajectory yields a higher observed solve rate than
uniform weighting
(\cref{fig:reflection-weighting-performance}). These findings motivate
future work to trace how retrospection gradients reshape action-relevant
representations and to test the role of this pathway in learning speed
and generalization.



\subsection{Ablating the Retrospection Learning Signal}
\input{overleaf_rr_public/assets/ablations/metrics}
\input{overleaf_rr_public/assets/nine_b/metrics}

\begin{figure}[htbp]
    \centering
    \vspace{-0.2in}
    \begin{subfigure}[t]{0.32\textwidth}
        \includegraphics[width=\linewidth]{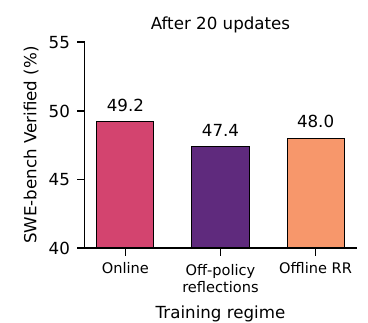}
        \caption{Online, off-policy, and offline.}
        \label{fig:ablation-policy}
    \end{subfigure}\hfill%
    \begin{subfigure}[t]{0.32\textwidth}
        \includegraphics[width=\linewidth]{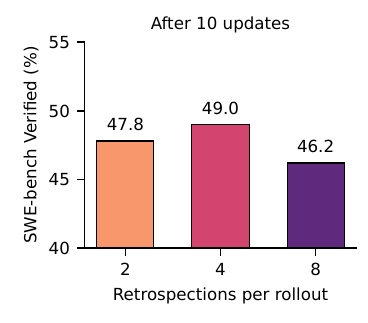}
        \caption{Retrospections per rollout.}
        \label{fig:ablation-reflections}
    \end{subfigure}\hfill%
    \begin{subfigure}[t]{0.32\textwidth}
        \includegraphics[width=\linewidth]{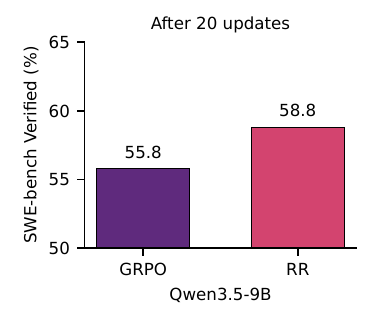}
        \caption{Qwen3.5-9B at 20 updates.}
        \label{fig:ablation-nine-b}
    \end{subfigure}
    \caption{
    \textbf{(a) Online RR has the highest observed solve rate.}
    SWE-bench Verified performance from Qwen3.5-4B with online,
    off-policy, or offline retrospection training. Off-policy reflections
    freeze the retrospection generator; offline RR freezes both the
    attempt and retrospection generators before training.
    \textbf{(b) Four retrospections per rollout perform best in this sweep.}
    Solve rates when varying retrospections per rollout and inversely
    varying source rollouts, holding the nominal retrospection batch
    fixed.
    \textbf{(c) The advantage over GRPO also holds for Qwen3.5-9B.}
    SWE-bench Verified solve rates under \methodabbr{} and GRPO from
    the same larger base model.
    Details in \cref{app:ablations}.}
    \vspace{-0.1in}
    \label{fig:ablations}
\end{figure}

\vspace{-0.05in}
\paragraph{Number of retrospections per rollout.}
Four retrospections per rollout give the highest observed test performance
in this sweep, suggesting a useful balance between rollout diversity
and retrospection diversity (\cref{fig:ablation-reflections}). We compare 2, 4, and 8 independently sampled
retrospections per rollout, using 128, 64, and 32 source rollouts,
respectively, to keep 256 nominal retrospection targets per update.
Fewer retrospections expose training to more source trajectories, whereas
more retrospections provide more independently sampled interpretations of
each trajectory. Neither extreme improves on the four-retrospection
control here. 
See \Cref{app:reflection_count} for details.

\vspace{-0.05in}
\paragraph{Different model.}
The advantage over GRPO also holds for Qwen3.5-9B:
after 20 updates, 
\methodabbr{} solves \NineBRRScore{} of SWE-bench
Verified versus \NineBGRPOScore{} for GRPO
(\cref{fig:ablation-nine-b}),
using \NineBRRHours{} versus
\NineBGRPOHours{} training hours. For Qwen3.5-9B, training performance
begins to plateau around 20 updates.
\Cref{app:nine_b} gives the training and evaluation protocol.

\vspace{-0.05in}
\paragraph{Verdict vs. no verdict.}
Providing the final correctness verdict does not improve the observed
SWE-bench Verified score in this comparison: after 20 updates, both
the verdict-conditioned baseline and the no-verdict variant achieve
\VerdictControlScore{} (\cref{fig:ablation-verdict}).
The no-verdict variant generates retrospections from the trajectory
summary and final patch, without the final grader feedback.
We hypothesize that environmental feedback already present in the
trajectory, including observations from the agent's own tests, provides
sufficient context to generate useful retrospections without an
explicit final correctness label.
\Cref{app:verdict_ablation} describes the training and evaluation protocols.

\vspace{-0.05in}
\paragraph{Off-policy reflections and offline.}
Does it matter whether the model learns from its own current retrospections?
\Cref{fig:ablation-policy} compares three training regimes at
 20 updates. Online RR, in which the evolving learner
generates both attempts and retrospections, achieves \PolicyOnlineScore{}.
Freezing the retrospection generator at the base model while continuing
to refresh learner attempts yields \PolicyOffPolicyScore{}.
Fully offline RR instead trains on a fixed corpus of base model attempts
and retrospections and reaches \PolicyOfflineScore{}.
The online procedure has the highest observed score, while fully offline
training outperforms off-policy reflections. This ordering is consistent
with a benefit from keeping the attempt and retrospection generators
aligned, rather than refreshing attempts alone.  See \Cref{app:policy_ablation} for details.

%% file: overleaf_rr_public/assets/frontier_zones/metrics.tex
\newcommand{\FrontierMasteredCount}{52}
\newcommand{\FrontierMasteredPct}{10.4\%}
\newcommand{\FrontierLearningCount}{309}
\newcommand{\FrontierLearningPct}{61.8\%}
\newcommand{\FrontierBeyondCount}{139}
\newcommand{\FrontierBeyondPct}{27.8\%}
\newcommand{\FrontierExecutionFailures}{35}

%% file: overleaf_rr_public/assets/training_efficiency/metrics.tex
\newcommand{\EfficiencyPadding}{0.45}
\newcommand{\EfficiencyRRCohorts}{93}
\newcommand{\EfficiencyRRHours}{4.32}
\newcommand{\EfficiencyRRSamples}{6,035}
\newcommand{\EfficiencyRRFinalReward}{0.483}
\newcommand{\EfficiencyRRPeakReward}{0.563}
\newcommand{\EfficiencyRRPeakUpdate}{18}
\newcommand{\EfficiencyGRPOCohorts}{42}
\newcommand{\EfficiencyGRPOHours}{8.30}
\newcommand{\EfficiencyGRPOSamples}{11,576}
\newcommand{\EfficiencyGRPOFinalReward}{0.522}
\newcommand{\EfficiencyGRPOPeakReward}{0.554}
\newcommand{\EfficiencyGRPOPeakUpdate}{23}
\newcommand{\EfficiencyTimeRatio}{1.92}

%% file: overleaf_rr_public/assets/combined_performance/metrics.tex
\newcommand{\CombinedGRPOScore}{47.8\%}
\newcommand{\CombinedRRScore}{49.2\%}
\newcommand{\CombinedCombinedScore}{50.4\%}

%% file: overleaf_rr_public/assets/hard_tasks/metrics.tex
\newcommand{\HardSympyFinal}{1.75\%}
\newcommand{\HardSympyPeak}{1.85\%}
\newcommand{\HardSymbiflowFinal}{3.29\%}
\newcommand{\HardSymbiflowPeak}{3.32\%}
\newcommand{\HardSymbiflowExtendedFinal}{4.49\%}
\newcommand{\HardSymbiflowExtendedPeak}{4.49\%}
\newcommand{\HardDjangoFinal}{0.73\%}
\newcommand{\HardDjangoPeak}{1.15\%}
\newcommand{\HardPylintEarlyFinal}{12.00\%}
\newcommand{\HardPylintEarlyPeak}{12.43\%}
\newcommand{\HardPylintFinal}{15.86\%}
\newcommand{\HardPylintPeak}{17.86\%}
\newcommand{\HardWindowMin}{1,024}
\newcommand{\HardWindowMax}{1,237}
\newcommand{\HardWindowPylintMax}{1,190}

%% file: overleaf_rr_public/assets/step_directions/metrics.tex
\newcommand{\StepGRPOMainCorrectIncreased}{44.14\%}
\newcommand{\StepGRPOMainIncorrectIncreased}{45.30\%}
\newcommand{\StepRRMainCorrectIncreased}{32.41\%}
\newcommand{\StepRRMainIncorrectIncreased}{21.21\%}
\newcommand{\StepGRPOCorrectIncreased}{24.57\%}
\newcommand{\StepGRPOCorrectMean}{-0.01384}
\newcommand{\StepGRPOIncorrectIncreased}{22.74\%}
\newcommand{\StepGRPOIncorrectMean}{-0.00713}
\newcommand{\StepGRPOMixedIncreased}{14.40\%}
\newcommand{\StepGRPOMixedMean}{-0.01159}
\newcommand{\StepGRPOUncertainIncreased}{17.39\%}
\newcommand{\StepGRPOUncertainMean}{-0.01153}
\newcommand{\StepRRCorrectIncreased}{21.36\%}
\newcommand{\StepRRCorrectMean}{-0.02831}
\newcommand{\StepRRIncorrectIncreased}{11.28\%}
\newcommand{\StepRRIncorrectMean}{-0.02042}
\newcommand{\StepRRMixedIncreased}{4.95\%}
\newcommand{\StepRRMixedMean}{-0.03552}
\newcommand{\StepRRUncertainIncreased}{6.52\%}
\newcommand{\StepRRUncertainMean}{-0.03379}
\newcommand{\StepGRPOMainGap}{-1.17}
\newcommand{\StepGRPOFinalGap}{1.83}
\newcommand{\StepRRMainGap}{11.20}
\newcommand{\StepRRFinalGap}{10.08}
\newcommand{\StepCorrectTrajectories}{253}
\newcommand{\StepCorrectTurns}{8,537}
\newcommand{\StepIncorrectTrajectories}{232}
\newcommand{\StepIncorrectTurns}{2,082}
\newcommand{\StepMixedTrajectories}{252}
\newcommand{\StepMixedTurns}{3,522}
\newcommand{\StepUncertainTrajectories}{46}
\newcommand{\StepUncertainTurns}{59}

%% file: overleaf_rr_public/assets/faster_prompt/metrics.tex
\newcommand{\FasterAllTokenReduction}{11.7\%}
\newcommand{\FasterAllTurnReduction}{13.1\%}
\newcommand{\FasterSolvedTokenReduction}{12.3\%}
\newcommand{\FasterSolvedTurnReduction}{12.9\%}
\newcommand{\FasterUnsolvedTokenReduction}{10.3\%}
\newcommand{\FasterUnsolvedTurnReduction}{13.0\%}
\newcommand{\FasterBaselinePassRate}{61.6\%}
\newcommand{\FasterBaselineTruncated}{25}
\newcommand{\FasterBaselineSolvedCount}{197}
\newcommand{\FasterBaselineUnsolvedCount}{123}
\newcommand{\FasterTreatmentPassRate}{58.4\%}
\newcommand{\FasterTreatmentTruncated}{16}
\newcommand{\FasterTreatmentSolvedCount}{187}
\newcommand{\FasterTreatmentUnsolvedCount}{133}
\newcommand{\FasterThroughTenTokenReduction}{10.3\%}
\newcommand{\FasterLateTokenChange}{-2.4\%}
\newcommand{\FasterThroughTenTurnReduction}{10.4\%}
\newcommand{\FasterLateTurnChange}{+3.2\%}

%% file: overleaf_rr_public/assets/ablations/metrics.tex
\newcommand{\ReflectionTwoScore}{47.8\%}
\newcommand{\ReflectionFourScore}{49.0\%}
\newcommand{\ReflectionEightScore}{46.2\%}
\newcommand{\ReflectionScoreSpread}{2.8}
\newcommand{\PolicyOnlineScore}{49.2\%}
\newcommand{\PolicyOffPolicyScore}{47.4\%}
\newcommand{\PolicyOfflineScore}{48.0\%}
\newcommand{\VerdictControlScore}{49.2\%}
\newcommand{\VerdictRemovedScore}{49.2\%}

%% file: overleaf_rr_public/assets/nine_b/metrics.tex
\newcommand{\NineBGRPOScore}{55.8\%}
\newcommand{\NineBRRScore}{58.8\%}
\newcommand{\NineBScoreGap}{3.0}
\newcommand{\NineBRRHours}{1.93}
\newcommand{\NineBRRSamples}{1,727}
\newcommand{\NineBGRPOHours}{5.33}
\newcommand{\NineBGRPOSamples}{7,152}

%% file: sections/related_works.tex
\section{Related Work}
\label{sec:related-work}
\vspace{-0.05in}

Work leveraging retrospection and feedback broadly follows three routes:
using interpretations to supervise task behavior, training on the feedback
itself, or retaining it as context for later attempts.
\emph{Retrospection reinforcement} asks whether \emph{learning to explain}
one's own experience can support \emph{learning to do}: improving subsequent
actions without directly training them or retaining explanations in context.
Our procedure uses supervised fine-tuning on retrospections rather than
reinforcement learning (RL). 
\Cref{fig:retrospection-tape-routes} illustrates these three routes and
distinguishes their conditioning context from their direct training targets.
An extended comparison is provided in \cref{app:related-work}.
\vspace{-0.05in}
\paragraph{Context internalization.}
Context internalization transfers privileged information into the model's
weights so that it acts as if that information were still in its prompt.
Training then
makes that behavior available without the context.
STaR illustrates a related supervised route: for a failed problem, it
supplies the correct answer as a hint, generates a rationale, and retains
the reasoning-and-answer sequence if the answer is correct. It then
fine-tunes on that sequence without the hint \citep{zelikman2022star}.
On-Policy Context Distillation (OPCD), RLTF-SD, and OPSD pursue a similar transfer using teacher predictions and,
in the latter two, reward-based updates
\citep{ye2026opcd,song2026rltf,penaloza2026privileged}.
ERL combines this idea with reflection-guided retries, distilling successful
revised behavior while also reinforcing attempts and reflections
\citep{shi2026erl}.
Simple self-distillation (SSD) removes the privileged context,
fine-tuning on the model's own samples at shifted temperatures \citep{zhang2026embarrassingly}. These methods share a common training target: \emph{actions}; our procedure instead makes the
\emph{retrospection itself} the training target.
\vspace{-0.05in}
\paragraph{Learning from textual feedback.}
Predicting feedback offers a more direct route from an interpretation to
the model's weights. RLTF-FM adds feedback prediction to reward-based task
training \citep{song2026rltf}; Early Experience trains reflections and
expert actions together \citep{zhang2025earlyexperience}.
Both therefore retain direct action supervision, although only RLTF uses RL.
Critique Fine-Tuning (CFT) is closer to our procedure: training on teacher critiques
can improve question answering without a separate answer-training objective
\citep{wang2025cft}. 
Self-Critique Fine-Tuning (SCFT) uses reattempts to filter critiques \citep{wang2026scft}.
We study this same transfer in an online agent setting, using
the actor's own outcome-conditioned retrospections rather than externally
taught or correctness-filtered critiques. We omit action-target losses
to isolate retrospection reinforcement.
\vspace{-0.05in}
\paragraph{Inference-time reflection.}
Alternatively, an interpretation can improve behavior by remaining
available to the actor. Reflexion and Self-Refine pass feedback into later
generations rather than update model weights
\citep{shinn2023reflexion,madaan2023selfrefine}.
Scattered Forest Search and Strategist similarly use feedback to search
over code or textual strategies \citep{light2025sfs,light2024strategist}.
Training the reflection generator does not remove this dependence:
Retroformer trains a reflector, and RetroAct trains both actor and
reflector, but their reflections still guide later trials
\citep{yao2023retroformer,feng2025retroact}.
Our procedure removes that route: later attempts receive no carried-over
reflection, so any benefit must transfer through the updated actor's
weights. \Cref{tab:related-work-comparison} summarizes these distinctions.

\begingroup
\setlength{\intextsep}{6pt}
\newcommand{\rwyes}{\ensuremath{\checkmark}}
\newcommand{\rwno}{\ensuremath{\times}}
\newcommand{\rwpartial}{\ensuremath{\sim}}
\begin{table}[t]
\centering
\caption{\footnotesize
\rwyes{} yes; \rwno{} no; \rwpartial{} partial:
---: no weight training.
\textbf{Actions}: directly trains task-solving outputs, including solutions
within critiques;
\textbf{Online}: fresh model generations during training; otherwise offline.
\textbf{Reflections}: reflection/critique tokens are training targets.
\textbf{Teacher/demo}: requires external teachers/demonstrations.
\textbf{Verdict}: requires verifier correctness judgments, not just environment observations.
\textbf{In context}: reflections used in context to affect actions.
\textbf{Reward}: scalar reward required for training.}
\vspace{-0.1in}
\label{tab:related-work-comparison}
\footnotesize
\setlength{\tabcolsep}{2pt}
\begin{tabularx}{\linewidth}{@{}l*{7}{>{\centering\arraybackslash}X}@{}}
\toprule
Method & \shortstack{Trains\\actions} & \shortstack{Online\\training} &
\shortstack{Trains\\reflections} & \shortstack{Teacher/\\demo} &
\shortstack{Verdict\\required} & \shortstack{Reflection\\in context} &
\shortstack{Scalar\\reward} \\
\midrule
RLTF & \rwyes & \rwyes & \rwyes & \rwyes & \rwyes & \rwyes & \rwyes \\
ERL & \rwyes & \rwyes & \rwyes & \rwno & \rwyes & \rwyes & \rwyes \\
OPCD (experience) & \rwyes & \rwyes & \rwno & \rwpartial & \rwno & \rwyes & \rwno \\
STaR & \rwyes & \rwyes & \rwno & \rwno & \rwyes & \rwno & \rwno \\
Early Experience SR & \rwyes & \rwno & \rwyes & \rwyes & \rwno & \rwyes & \rwno \\
Reflexion / Self-Refine & \rwno & --- & \rwno & \rwno & \rwno & \rwyes & \rwno \\
CFT & \rwno & \rwno & \rwyes & \rwyes & \rwyes & \rwno & \rwno \\
SCFT & \rwyes & \rwno & \rwyes & \rwno & \rwyes & \rwno & \rwno \\
\textbf{\shortstack[l]{\paradigmabbr (ours)}} & \rwno & \rwyes & \rwyes & \rwno & \rwno & \rwno & \rwno \\
\bottomrule
\end{tabularx}
\vspace{-0.2in}
\end{table}
\endgroup


%% file: sections/conclusion.tex
\vspace{-0.05in}
\section{Limitations}
\vspace{-0.05in}
\label{sec:limitations}
The cost of training and evaluating long-horizon agents limits the breadth of our
experiments, which focus on software engineering with Qwen3.5-4B. A single 40-update GRPO training run costs \$500, for example, at current market rates for GPUs, and a single evaluation run on SWE-bench Pro incurs \$200 in GPU costs alone. We chose
\methodabbr{} for its simplicity rather than optimality: even this minimal
procedure can improve subsequent actions by training only on self-generated
explanations. Our behavioral analyses support this transfer and motivate
larger, controlled studies across models and domains to identify its
causal mechanisms and guide more effective ways to generate and
internalize useful explanations.

\vspace{-0.05in}
\section{Conclusion}
\vspace{-0.05in}
\label{sec:conclusion}

We explore whether an agent can improve its future actions by training only on self-generated retrospective explanations of its own experience. \methodabbr{} isolates this question with a minimal online procedure in which effects transfer to later attempts through the updated weights, and our findings show that learning to explain can also improve learning to do. This motivates algorithms that learn to assign credit and shape
behavior through language, rather than encoding every desired change in
a scalar reward. This direction is especially relevant for long-horizon
agents, where sparse outcomes provide little guidance about which
intermediate decisions mattered. Grounding retrospections in observations
could also broaden learning beyond curated tasks with human-crafted
verifiers, making ordinary interaction a source of supervision. The next
challenge is to learn which interpretations deserve to be internalized,
so that agents improve from evidence rather than reinforce plausible but
mistaken explanations.

%% file: sections/statements.tex
\subsection*{AI use statement}
We used generative AI tools, including GitHub Copilot, to assist with
drafting and revising the paper, literature retrieval and discovery,
research ideation and execution, and the development of mathematical
claims and proofs. Writing assistance included proposing and refining
the title, abstract, submission summary, and keywords. The authors have
checked all AI-assisted text, references, analyses, and mathematical
claims. We take responsibility for the final content of this work,
including text, claims, and artifacts produced with generative-AI
assistance. Separately, the model-generated retrospective training targets
are part of the experimental method described in
\cref{sec:retrospection}. GPT-6 Astra also supplied trajectory-turn
correctness labels and semantic retrospection-span annotations for the
behavioral analyses in \cref{app:behavioral_induction}; these annotations
are model judgments, not independently audited ground truth.

\subsection*{Ethics statement}
This work studies software-engineering agents on existing benchmarks
using model-generated attempts and retrospections, rather than deploying
them to end users. Improved coding agents can support software maintenance,
but can also generate insecure or incorrect code and lower barriers to
misuse. Benchmark test success does not establish safety, security, or
reliability outside the evaluated tasks. Self-generated retrospections
can reinforce mistaken explanations or biased assumptions and should
not be treated as faithful accounts of a model's internal reasoning.
Deployment would require independent validation, restricted tool
permissions, isolated execution environments, and human oversight
appropriate to the application. Reuse or redistribution of the underlying
models, benchmark data, repository code, and derived artifacts remains
subject to their respective licenses and terms; public availability
does not by itself remove privacy or intellectual-property concerns.
Training and evaluation also incur computational and environmental costs.
Our reported efficiency improvements are specific to the measured
experimental budgets, not estimates of lifecycle environmental impact.

\subsection*{Reproducibility statement}
The training objective and online update procedure are specified in
\cref{sec:retrospection}. \Cref{app:experimental_setup} documents the
base model, training tasks, agent environment, sampling settings,
target preprocessing and masking, optimizer hyperparameters, compute,
and checkpointing. Complete retrospection prompts and generated examples
appear in \cref{app:retrospection_examples}. Baseline implementations
are described in \cref{sec:baselines}, and experiment-specific protocols,
budgets, evaluation settings, aggregation, and uncertainty estimates
are reported in
\cref{app:generalization_efficiency,app:beyond_frontier_learning,app:behavioral_induction,app:ablations}.
The paper-source assets retain plotting scripts and saved numerical
inputs for regenerating figures without repeating model training or
inference; plot regeneration is distinct from reproducing the underlying
experiments. We will release selected components of the training and
evaluation code upon acceptance. Exact reruns may differ because
generation is stochastic and asynchronous execution affects batch
composition, as described in \cref{app:experimental_setup}.



%% file: sections/appendix/related_works_extended.tex
\section{Related Work (Extended)}
\label{app:related-work}

Retrospection reinforcement (RR) studies whether \emph{learning to explain}
past experience can improve future actions. Our procedure collects the
agent's own attempts and available feedback, generates retrospections with
the same model, and applies next-token cross-entropy only to the
retrospective continuations. Subsequent attempts use updated weights
without the earlier retrospections in context. The relevant distinction
from prior work is therefore not the presence of reflective language,
but its role in learning: feedback can supervise task behavior, become
a prediction target itself, or remain available as inference-time
context. Following the main text, we organize the comparison around
these three routes, illustrated in \cref{fig:retrospection-tape-routes}.

\input{overleaf_rr_public/assets/rr_ar_concepts/figure_routes_tapes}

\subsection{Context internalization: learning behavior informed by feedback}

An interpretation can improve a policy by helping construct better task
outputs for training. Imitation learning from language feedback (ILF)
generates feedback-conditioned revisions and trains the task model on
selected revisions rather than on the feedback itself
\citep{scheurer2023scale,chen2023codefeedback}. STaR similarly fine-tunes
on generated reasoning-and-answer sequences that yield correct answers
\citep{zelikman2022star}. Failed problems can be revisited by supplying
the correct answer as a hint and generating a rationalization, so STaR
does not simply discard unsuccessful experience. In both cases,
however, the retained target practices a successful response. RR instead
trains an account of the completed attempt, without requiring a successful
replacement trajectory.

Context distillation makes the transfer from guidance to behavior
explicit: a teacher receives information that the deployed student will
not receive, and training aligns the student's predictions with the
teacher's \citep{askell2021general,snell2022distilling}. On-Policy Context
Distillation (OPCD) applies this principle along student-generated
responses \citep{ye2026opcd}. Its experiential variant extracts reusable
lessons from trajectories and supplies them to the teacher during
distillation. Online Experiential Learning (OEL) repeats experience
collection, extraction, and consolidation \citep{ye2026oel}. These methods
already pursue online learning from experience without retaining the
lessons at deployment. Their direct targets are nevertheless
\emph{task-response distributions}, whereas RR predicts the
\emph{retrospective explanation}.

Feedback-conditioned self-distillation provides closely related
comparisons. Self-Distillation Policy Optimization (SDPO) uses an informed
self-teacher to supervise response-token distributions; rich feedback
can supply this signal without a complete corrected response
\citep{hubotter2026sdpo}. Reflection-Enhanced Self-Distillation (RESD)
adds failure diagnoses and accumulated guidance to the teacher's context
\citep{zhang2026resd}, while Procedural Memory Distillation (PMD)
internalizes experience through memory-conditioned behavioral
distillation \citep{liu2026pmd}. HERO makes the target distinction
particularly clear: hindsight reflections condition a teacher that
supervises the original assistant-action tokens, not the reflections
\citep{liu2026hero}. Thus self-generated lessons, learning from failure,
and memory-free deployment do not by themselves distinguish RR.
The difference is whether the lesson informs an action-training signal
or is itself the supervised continuation.

Experiential Reinforcement Learning (ERL) combines reflection-guided
retries with reward-based updates to attempts and reflections
\citep{shi2026erl}. It also internalizes positive-reward revisions under
the original task prompt, enabling deployment without the reflection.
RLTF's Self-Distillation variant likewise transfers feedback-guided
second-turn behavior to the original-prompt policy alongside multi-turn
RL \citep{song2026rltf}. Privileged-information distillation combines
reward and distribution-alignment objectives using training-only
guidance \citep{penaloza2026privileged}. These methods directly optimize
task behavior; our procedure omits both reward-driven action updates
and revised-response imitation. This is a distinction between training
targets, not between methods with and without RL: supervised imitation
also trains behavior, and RL can train reflections.

\subsection{Learning from textual feedback: predicting the interpretation}
\label{app:feedback-targets}

Training on evaluative language offers a more direct precedent for RR.
Early dialogue systems learn to predict a teacher's response after an
attempted answer, with shared parameters allowing feedback prediction
to improve answering without the feedback-prediction component at
evaluation \citep{weston2016dialog,li2016dialogue}. These studies establish
that transfer from post-answer language to task performance predates
current reflection methods. Their supervision is externally supplied,
rather than generated by an agent interpreting its own interaction.

RLTF's Feedback Modeling variant applies cross-entropy to feedback
tokens, treating the sampled answer as fixed, alongside reward-based
task training \citep{song2026rltf}. It is therefore a close precedent
for our loss-target choice, distinct from RLTF's Self-Distillation
variant discussed above. Agent Learning via Early Experience also
uses observed transitions as richer learning material than scalar
rewards: its self-reflection route jointly predicts a reflection and an
expert action \citep{zhang2025earlyexperience}. Both retain direct task
training. RR removes that component to isolate transfer from
retrospection prediction to subsequent actions.

Critique Fine-Tuning (CFT) is closer still. It supervises critiques
conditioned on questions and candidate solutions, then evaluates the
model's ability to answer questions directly \citep{wang2025cft}.
Its main targets come from an external critique teacher, and its
experiments also consider candidate answers generated by the student.
CFT thus already demonstrates the central possibility that learning
to evaluate an answer can improve generation without a separate
answer-training objective. Our study extends this question to online
agent experience, using retrospections authored by the continually
updated actor rather than a fixed collection of externally taught
critiques.

Self-Critique Fine-Tuning (SCFT) further narrows this distinction
\citep{wang2026scft}. The same model generates solutions and critiques,
and ground-truth-based filtering selects acceptable critiques for
supervised fine-tuning. Standalone SCFT supports direct answering
without a supplied critique; its additional RLERR stage is a separate
RL component. Neither self-generated critique targets nor
critique-to-answer transfer is therefore unique to RR. Moreover, SCFT
targets can include corrected derivations and final answers: masking
the input solution does not make the supervised content solution-free.
Our procedure trains on self-generated post-attempt accounts, without
requiring a successful revised solution as a training target. These
accounts may summarize events, track changes in beliefs, or discuss
evidence and corrective actions. The distinction concerns the training
task, not an absence of actionable information.

The remaining differences concern the learning setting and evidence.
RR refreshes attempts and retrospective targets as the policy changes,
conditions generation on the observed trajectory and available outcome
feedback, and admits both successful and unsuccessful attempts. It
does not independently verify the explanation's correctness. A reliable
test verdict can ground a retrospection without validating its causal
claims, so omitting critique-quality filtering is a limitation as well
as a simplification. The contribution is to isolate this online,
self-authored, retrospection-only configuration, not to introduce
critique prediction as a new objective.

\subsection{Inference-time reflection: retaining interpretations as context}

Reflexion stores verbal self-reflections in episodic memory and supplies
them to subsequent attempts \citep{shinn2023reflexion}. Self-Refine uses
feedback from the same model to revise an output without additional
weight training \citep{madaan2023selfrefine}. ExpeL extracts reusable
insights and retrieves experience for future tasks
\citep{zhao2023expel}. These methods show that reflection can support
both local correction and cross-task improvement through an external
memory. The contrast with RR is not whether learning persists across
tasks, but whether the actor must receive the retained text.
Scattered Forest Search and Strategist similarly use feedback to
guide search over code or textual strategies rather than train the
actor on retrospective continuations
\citep{light2025sfs,light2024strategist}.
Related multi-agent work includes AvalonBench, which evaluates LLMs
in the social-deduction game Avalon \citep{light2023avalonbench},
and PIANIST, which generates world models with LLMs for Monte Carlo
tree search \citep{light2024pianist}.

Learning the reflection generator does not necessarily remove this
inference-time dependence. Retroformer trains a reflector to produce
feedback that helps a frozen actor; RetroAct jointly trains planning
and reflection, including a shared-model variant
\citep{yao2023retroformer,feng2025retroact}. Their reflections still
guide later trials. The relevant distinction is consequently not
separate versus shared weights, but the pathway through which a
reflection changes behavior. RR omits carried-over retrospections and
an added reflection step on subsequent tasks, testing whether the
effect transfers through the actor's updated parameters.

\paragraph{Inference-time search and sampling.}
Beyond reflection, inference-time methods improve solution generation
by changing how computation is allocated. DISC adaptively decomposes
solution traces and prioritizes difficult steps \citep{light2025disc},
while diversified sampling uses prompt perturbations to broaden
candidate exploration \citep{wang2025divsampling}.
These approaches modify inference-time search and selection rather
than train an actor to predict retrospective explanations.

\paragraph{Scope of the comparison.}
Retrospection-only training isolates \emph{learning to explain,
evaluated by doing}. Unlike reward-based updates such as GRPO
\citep{shao2024deepseekmath}, its loss does not weight attempted
actions by their outcomes. RLVR often relies on careful curation of
problems within the model's learning zone, either manually or through
automated curricula such as Actor-Curator \citep{gu2026actorcurator}.
By contrast, retrospective targets remain available
even when all sampled attempts fail and a GRPO group has no relative
outcome advantage. These properties motivate future work on generating
useful explanations from failed attempts and identifying the mechanisms
and conditions that make transfer through shared parameters reliable.
The relevant evidence is improved behavior on
independent attempts without retained retrospections, not explanation
fluency alone. The closest prior work motivates comparisons with
critique training, feedback-conditioned action distillation, and
retained textual memory, rather than a general claim of superiority
over those routes.

%% file: overleaf_rr_public/assets/rr_ar_concepts/figure_routes_tapes.tex
\begin{figure}[ht]
    \centering
    \includegraphics[width=\linewidth]{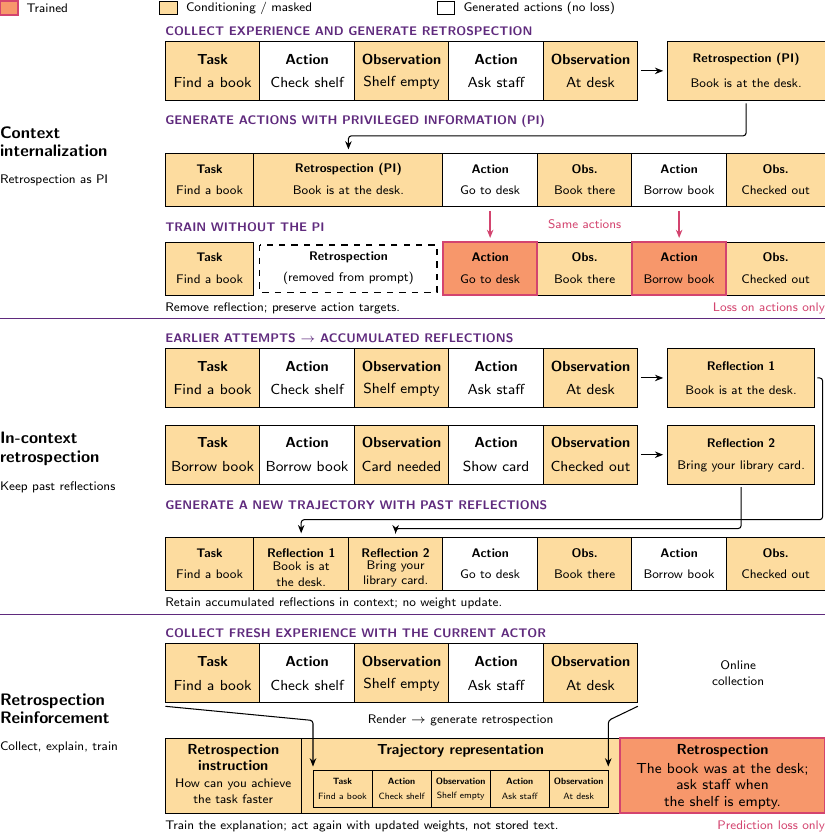}
    \captionsetup{font=footnotesize,skip=3pt}
    \caption{\textbf{Three uses of retrospection.}
    \textbf{Top:} context internalization collects experience, generates a
    retrospection containing privileged information (PI), and uses it to
    generate action targets. Training removes that context (dashed box).
    \textbf{Middle:} earlier attempts yield reflections accumulated in later
    prompts, without a weight update. \textbf{Bottom:} RR collects fresh
    experience and trains its retrospection; later attempts use updated weights
    without stored text. Black arrows show data flow, not gradient updates.
    Orange marks training targets; yellow marks conditioning context.
    Miniatures are trajectory representations, which may be rendered,
    summarized, or truncated.
    These illustrative routes can be combined; the top row depicts
    action-target distillation, not every internalization method.}
    \label{fig:retrospection-tape-routes}
\end{figure}

%% file: sections/appendix/experimental_setup.tex
\section{Experimental Setup}
\label{app:experimental_setup}

This section describes the Qwen3.5-4B online training procedure for
\methodabbr{}. Training budgets and experiment-specific changes are
reported in \cref{app:generalization_efficiency,app:beyond_frontier_learning,app:behavioral_induction,app:ablations}.

\subsection{Model, tasks, and agent environment}

We initialize the policy from Qwen3.5-4B
\citep{qwen2026qwen35fourb} and train on SWE-rebench-767, a fixed
collection of 767 software-engineering tasks. Each task is executed in
a repository-specific container. The
agent can read, write, and edit files, search the repository, and execute
shell commands. Its system instruction is:
\begin{quote}\small
You are a software-engineering agent working in a checked-out repository.
Use the available tools to inspect files, make focused edits, and run commands.
Solve the task, validate when practical, then give a brief final answer.
\end{quote}
The environment supplies the task description and evaluates the resulting
patch. We allow at most 100 agent turns, 8,192 generated tokens per turn,
and a context window of 131,072 tokens. Context compaction is disabled.
The solver samples at temperature 1.0 with top-$p=1.0$.

Task order is shuffled and the dataset is traversed cyclically, with
reshuffling at each epoch. Because generation and environment execution
are asynchronous, completion order also influences batch composition.

\subsection{Retrospection generation and supervision}
\label{app:retrospection_generation}

After each attempted solution, the same policy generates four independent
retrospections, conditioned on the same task, trajectory summary, patch,
and test feedback. The default prompt used in these experiments requests
a \emph{short paragraph without lists} that identifies a consequential
assumption or decision, relates it to supporting or contradicting evidence, gives a correction when
appropriate, and states concrete triggers for applying the lesson.
It explicitly distinguishes the correctness of individual decisions
from the overall verdict and asks the model to acknowledge insufficient
evidence. The complete prompt and two generated examples are provided in
\cref{app:retrospection_examples}.

We sample retrospections at temperature 0.9 with top-$p=1.0$ and an
8,192-token output limit. The short-paragraph instruction specifies this
prompt's requested answer format, not the general form of retrospection,
and does not exclude the model's preceding thinking from supervision.
Both failed and successful attempts contribute
targets, without reward weighting or a semantic-quality filter.

The prompt reports the raw test outcome separately from reward adjustments
and the reason for termination. An attempt that passes its tests but
exhausts the turn budget therefore retains a passing test verdict;
budget exhaustion is reported separately. If raw test feedback is
unavailable, the verdict is marked unknown rather than inferred from
the shaped reward.

The no-verdict ablation removes the entire post-attempt verdict and
test-feedback block from the retrospection prompt, rather than only
the pass/fail label. Observations recorded during the task attempt
remain available, including any test output already present in that
trajectory.
The faster-retrospection experiment instead retains the verdict and uses
it to choose the instruction (\cref{app:faster_rollouts}).

The retrospective context is constructed from a bounded rendering of
the interaction, rather than the complete solver context.
\Cref{tab:retrospection_context} lists the character budgets.
For fields exceeding a budget, we retain the first two-thirds and last
one-third, separated by a truncation marker. Thought, action, and
observation fields are shortened independently before the trajectory-level
budget is applied. The full prompt and target must also fit the
131,072-token training sequence limit.

\begin{table}[H]
\centering
\small
\caption{Character budgets for constructing the retrospection input.
Budgets exclude inserted truncation markers and structural headings.}
\label{tab:retrospection_context}
\begin{tabularx}{0.85\linewidth}{@{}Xr@{}}
\toprule
Input component & Retained characters \\
\midrule
Task description & 4,000 \\
Thought, action, or observation within each step & 600 each \\
Concatenated trajectory summary & 24,000 \\
Final patch & 2,000 \\
Test output & 3,000 \\
\bottomrule
\end{tabularx}
\end{table}

We discard generations that terminate at the output-length limit,
empty or failed generations, and prompt--target pairs that exceed the
training sequence limit. Preprocessing removes surrounding whitespace,
trailing termination markers, and a leading empty thinking block, while
preserving nonempty thinking and the final answer. An end-of-turn token
and newline are appended to each accepted target. All target tokens are
supervised, including thinking and the end-of-turn suffix; all prompt
tokens are masked. Rejected targets are not replaced with action-token
supervision.

To detect degenerate outputs, training stops after two consecutive batches
in which the median target length falls below 100 tokens or more than
10\% of targets contain at most 10 tokens. No minimum-length filter
is applied to individual retrospections.

\subsection{Online optimization}
\label{app:roft_loss_implementation}

Each nominal update uses 64 source attempts and four retrospections
per attempt, yielding up to
256 targets before rejection. We make one optimization pass through
the retained targets and do not replay earlier batches. Source groups
are consumed in completion order. We discard groups that began generation
with weights more than one update behind the current policy.
Serving weights are refreshed after each optimizer update, so a
multi-turn attempt and its subsequent retrospections need not use a
single frozen policy snapshot.

The objective in \cref{eq:ro} averages over all supervised
tokens in the retained batch.
We optimize all model parameters with AdamW, a constant learning rate
of $10^{-6}$, and no warmup. The data gradient is scaled by a constant
factor of four before norm clipping.
Relative token weighting is unchanged.
We use neither a reference-policy KL penalty nor an entropy bonus.
Optimizer settings and generation hyperparameters are summarized in
\cref{tab:roft_hyperparameters}.

\begin{table}[H]
\centering
\small
\caption{Reference online training and generation hyperparameters for
\methodabbr{}. Experiment-specific settings take precedence. Batch sizes
are nominal; targets rejected during preprocessing incur no loss.}
\label{tab:roft_hyperparameters}
\begin{tabularx}{\linewidth}{@{}p{0.48\linewidth}X@{}}
\toprule
Hyperparameter & Value \\
\midrule
Initial model & Qwen3.5-4B \\
Training tasks & 767 \\
Optimizer updates & Specified per experiment \\
Source attempts per update & 64 \\
Solver attempts per task selection & 1 \\
Retrospections per source & 4 \\
Target batch / microbatch size & 256 / 1 \\
Optimization passes per batch & 1; no replay \\
Optimizer & AdamW \\
Learning rate / schedule / warmup & $10^{-6}$ / constant / none \\
Adam $(\beta_1,\beta_2)$ / $\epsilon$ & $(0.9,0.98)$ / $10^{-8}$ \\
Weight decay / gradient-norm clip & 0.1 / 1.0 \\
Loss reduction & Mean over retained target tokens \\
Reference-KL / entropy coefficients & 0 / 0 \\
Solver temperature / top-$p$ & 1.0 / 1.0 \\
Retrospection temperature / top-$p$ & 0.9 / 1.0 \\
Top-$k$ restriction & None \\
Maximum solver turns & 100 \\
Maximum generated tokens per solver turn & 8,192 \\
Maximum generated tokens per retrospection & 8,192 \\
Solver context / total response limit & 131,072 / 131,072 tokens \\
Training sequence limit & 131,072 tokens \\
Context compaction & Disabled \\
Maximum policy lag at generation start & One update \\
Training precision / gradient accumulation & BF16 / FP32 \\
Attention / hidden dropout & 0 / 0 \\
\bottomrule
\end{tabularx}
\end{table}

\subsection{Compute and checkpointing}

We use eight NVIDIA B200 GPUs: four for training and four for
generating solution attempts and retrospections. Training uses
four-way context parallelism and accumulates gradients over the
complete update.

We save checkpoints every five successful optimizer updates and at the
final update. Evaluated checkpoints are reported in
\cref{app:generalization_efficiency,app:single_task_test_performance,app:ablations}.
Held-out evaluations are conducted separately from the online training
loop, with no stored retrospection or added reflection step in the
solver context.

%% file: sections/appendix/baselines.tex
\section{Baselines}
\label{sec:baselines}

We compare \methodabbr{} with baselines spanning reinforcement learning
(GRPO), distillation from teacher-generated explanations (CFT), context
internalization (ERL), and self-critique-based revision (SCFT). Our GRPO
baseline uses an enhanced variant previously shown to improve performance.

\subsection{Held-out performance}
\label{app:baseline_results}

\begin{figure}[htbp]
    \centering
    \includegraphics[width=0.80\linewidth]{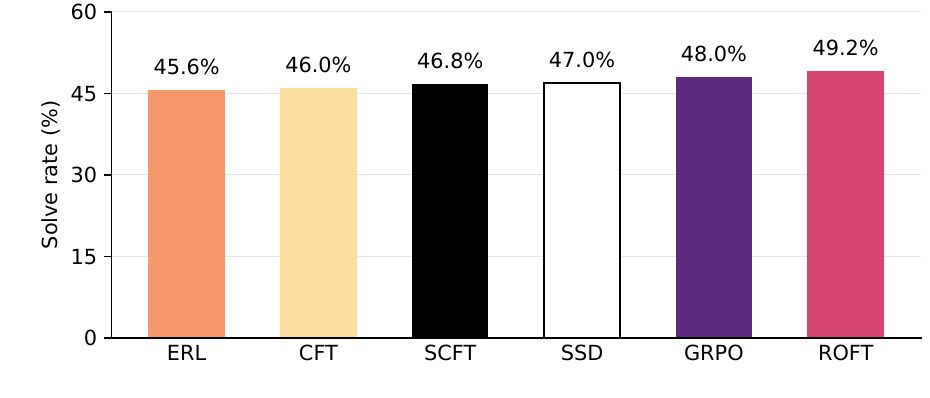}
    \caption{SWE-bench Verified solve rates for Qwen3.5-4B}
    \label{fig:prior-work-baselines}
\end{figure}

\input{overleaf_rr_public/assets/prior_work_baselines/metrics}
On SWE-bench Verified, ERL, CFT, SCFT, SSD, GRPO, and \methodabbr{} achieve solve rates
of \PriorBaselineERLScore{}, \PriorBaselineCFTScore{},
\PriorBaselineSCFTScore{}, \PriorBaselineSSDScore{}, \PriorBaselineGRPOScore{}, and
\PriorBaselineROFTScore{}, respectively (\cref{fig:prior-work-baselines}).

We hypothesize that off-policy supervision contributes to the lower solve
rates of ERL and CFT relative to \methodabbr{}. In ERL, successful reattempts
are generated with a reflection that conveys privileged information, but
this reflection is removed during context internalization, creating a
mismatch between the contexts used to generate and train on the reattempt.
In CFT, the critique targets are generated by a separate teacher model
rather than the learner itself. This interpretation is consistent with our
\methodabbr{} ablations: generating retrospections with a frozen base model
yields lower held-out performance than using the evolving learner
(\cref{fig:ablation-policy,app:policy_ablation}). 

The comparison with SCFT suggests that supervising successful
revision trajectories does not necessarily improve
first-attempt performance as much as training on retrospections of the
initial attempt.

\subsection{GRPO}
\label{app:grpo}

We use the same GRPO setup that achieved competitive performance in
FrogNano \citep{kim2026frognano}.

Group Relative Policy Optimization (GRPO; \citealp{shao2024deepseekmath})
learns from differences in the outcomes of multiple attempts at the same
task, without a learned value function. For each task $x$, we sample a
group of $G=8$ trajectories $\{\tau_i\}_{i=1}^{G}$ and compute a scalar
reward $R_i$ for each. The group-relative advantage is
\begin{equation}
    A_i = \frac{R_i-\bar R}{s_R+10^{-6}},
    \qquad
    \bar R = \frac{1}{G}\sum_{j=1}^{G}R_j,
    \qquad
    s_R^2 = \frac{1}{G-1}\sum_{j=1}^{G}(R_j-\bar R)^2.
    \label{eq:grpo_advantage}
\end{equation}
The same advantage is assigned to every assistant-generated token in
$\tau_i$, including reasoning and tool calls. Task instructions and
environment observations provide context but incur no loss. Thus,
unlike \methodabbr{}, this baseline directly reinforces task-solving
behavior rather than a subsequent retrospection.

\paragraph{Policy update.}
We retain GRPO's advantage estimator but replace its PPO-clipped surrogate
with DPPO--Binary-TV, the binary total-variation variant of
\citet{qi2026dppo}. This controls absolute probability changes rather than
ratios, avoiding the disproportionately tight constraints that ratio
clipping imposes on low-probability tokens. This extended baseline is
denoted GRPO in our comparisons. Let
$p_{it}=\pi_\theta(a_{it}\mid h_{it})$ be the probability of assistant
token $a_{it}$ given its interaction history $h_{it}$, and let $q_{it}$
be its recorded sampling probability. With
$\rho_{it}=p_{it}/q_{it}$, we maximize
\begin{equation}
    J(\theta)
    = \mathbb{E}_{\text{accepted groups}}
      \left[
      \frac{1}{G}\sum_{i=1}^{G}\frac{1}{T_i}
      \sum_{t=1}^{T_i}m_{it}\rho_{it}A_i
      \right],
    \qquad
    m_{it} =
    \begin{cases}
        \mathbf{1}[p_{it}-q_{it}\leq 0.2], & A_i>0,\\
        \mathbf{1}[p_{it}-q_{it}\geq -0.2], & A_i\leq 0,
    \end{cases}
    \label{eq:grpo_update}
\end{equation}
where $T_i$ counts assistant-generated tokens. Sampling probabilities,
advantages, and masks are held fixed during differentiation. The mask
suppresses contributions once a token's probability has moved more than
0.2 in the advantage-favored direction, following the Binary-TV threshold
used by \citet{qi2026dppo}. We retain GRPO's within-trajectory averaging
followed by averaging across trajectories, without renormalizing after
masking. Following the KL-free formulation of DAPO
\citep{yu2025dapo}, we omit the reference-policy penalty to permit
adaptation away from the initial policy. The objective includes no
additional KL regularizer or entropy bonus.

\paragraph{Reward and group selection.}
The reward augments binary test success with task-specific shaping.
Correctness-conditioned length control \citep{aggarwal2025l1} motivates
preferring concise successful attempts without rewarding short failures.
A successful, normally completed attempt receives
$R_i=1-0.1\max\{0,\log(N_i/8192)\}$, where $N_i$ is its total number of
generated assistant tokens. The logarithmic penalty therefore applies
only to successful completions exceeding 8,192 tokens. DAPO identifies
reward noise from treating truncation as failure \citep{yu2025dapo};
we instead give test-passing truncated attempts partial credit of $0.5$,
preserving evidence of correctness while penalizing budget exhaustion.
Failed attempts and other non-completed attempts receive zero.
The logarithmic schedule and partial-credit value are our task-specific
settings, not the reward functions proposed in those works.

Following DAPO's dynamic sampling \citep{yu2025dapo}, we discard
zero-advantage groups and replenish the batch to avoid spending optimizer
updates on uninformative samples. Here this criterion is applied to
\emph{shaped} rewards: we accept 32 groups (256 trajectories) whose
rewards are not all identical. An all-failure group therefore provides
no update, whereas an all-success group can remain informative through
differences in length or termination.

\paragraph{Training protocol.}
For the main Qwen3.5-4B comparison, we use the same initialization,
SWE-rebench-767 tasks, and agent environment as \methodabbr{}
(\cref{app:experimental_setup}). Each accepted batch receives one
AdamW update at a constant learning rate of $10^{-6}$, without replay.
We overlap rollout generation and training to reduce waiting for
variable-duration attempts, with bounded staleness as in asynchronous
RL systems such as AReaL \citep{fu2025areal}. We discard groups more
than three policy updates old. The importance ratios in
\cref{eq:grpo_update} use recorded token-level sampling probabilities,
anchoring the update to the behavior policy as advocated by
\citet{qi2026dppo}, rather than assuming a single frozen rollout policy.
Sampling uses temperature 1.0 and top-$p=1.0$, with at most 100 agent
turns and a 131,072-token context window without compaction.
Training-cost accounting includes rejected attempts, and held-out
evaluation reports unshaped task success
(\cref{app:generalization_efficiency}).

\subsection{ERL}
\label{app:erl}

Experiential Reinforcement Learning (ERL; \citealp{shi2026erl}) uses
reflection-guided retries to learn from failed attempts and internalize
successful revisions. In our adaptation, the policy first attempts a
task without additional guidance. If the attempt fails, the same policy
generates a reflection conditioned on the task, the failed trajectory,
and the environment verdict. It then reattempts the task conditioned
on the task and reflection, without directly including the first
trajectory or its verdict in the retry context.

We assign the second attempt's binary task-success reward to both the
reflection and the revised attempt. This yields a success-guided
 objective: successful retries provide training rewards for
both generations, whereas failed retries contribute no loss. The
reflection is trained under its original retrospective context.
Crucially, we remove the reflection when training on the successful
retry, retaining the original task prompt and the retry's own
interaction history. This internalizes reflection-guided behavior
without requiring the reflection at inference time. Only
assistant-generated target tokens incur loss; conditioning inputs and
environment observations are masked. Unlike \methodabbr{}, which trains
on retrospections without requiring a successful revision, this
baseline uses downstream task success to select reflections and
directly supervises revised task-solving behavior.

\subsection{CFT}
\label{app:cft}

Critique Fine-Tuning (CFT; \citealp{wang2025cft}) transfers knowledge
from a teacher to a student by supervising critiques of candidate
solutions rather than the solutions themselves. The student learns to
generate the teacher's critique conditioned on the question and candidate
solution, with the aim of improving subsequent task performance without
requiring critiques at inference time.

We adapt CFT to online agentic learning using \texttt{gpt-6-luna} as
the critique teacher and Qwen3.5-4B as the student. The evolving student
collects coding trajectories, and the teacher generates four independent
critiques per attempt, conditioned on the task, recorded interaction,
and environment verdict. 
Each update uses up to 256
critiques from 64 source attempts. We minimize next-token cross-entropy
averaged over supervised target tokens; only the teacher's returned
critique and its termination tokens incur loss, while all conditioning
inputs are masked.

\subsection{SCFT}
\label{app:scft}

Self-Critique Fine-Tuning (SCFT; \citealp{wang2026scft}) trains a model
on successful revisions of its own solutions, without an external
critique teacher. The model first attempts a task, then critiques and
revises its solution conditioned on the original attempt, without being
told whether that attempt was correct. Revisions are retained only if
their final solutions pass the task's correctness criterion, irrespective
of the first attempt's outcome. Supervised fine-tuning then maximizes
the likelihood of the complete critique-and-revision continuation,
with the initial attempt serving only as context. The procedure
aims to transfer learned self-correction to ordinary
first-attempt solving.

In our agentic adaptation, a frozen Qwen3.5-4B generates two initial
attempts per training task and four independent critique-and-revision
branches per attempt. 
Training uses 256 accepted
revision trajectories per update, with cross-entropy averaged over
all supervised target tokens. Only second-stage assistant tokens,
including critique reasoning, tool calls, and the final response,
incur loss; the original attempt, conditioning instructions, and tool
observations are masked. 

\subsection{SSD}
\label{app:ssd}

Simple Self-Distillation (SSD; \citealp{zhang2026embarrassingly})
fine-tunes a model on its own sampled solutions, without a stronger
teacher or correctness-based selection. A frozen model generates an
offline corpus, and a student initialized from the same weights learns
to reproduce these samples through next-token cross-entropy. Successful
and unsuccessful attempts both provide targets.

We adapt this procedure to agentic coding using trajectories collected
by the original Qwen3.5-4B on the training tasks, with two attempts per
task sampled at temperature 1.0 and top-$p=1.0$. Each update
uses 256 complete trajectories, with cross-entropy averaged over all
supervised target tokens. 
We supervise the original assistant-generated
reasoning, tool calls, and final response, including assistant termination
tokens supplied by the harness; task instructions and tool
observations provide context but incur no loss. 
Thus, unlike \methodabbr{}, SSD directly imitates task-solving behavior,
including unsuccessful behavior, rather than learning to explain it.

%% file: overleaf_rr_public/assets/prior_work_baselines/metrics.tex
\newcommand{\PriorBaselineERLScore}{45.6\%}
\newcommand{\PriorBaselineCFTScore}{46.0\%}
\newcommand{\PriorBaselineSCFTScore}{46.8\%}
\newcommand{\PriorBaselineSSDScore}{47.0\%}
\newcommand{\PriorBaselineGRPOScore}{48.0\%}
\newcommand{\PriorBaselineROFTScore}{49.2\%}

%% file: overleaf_rr_public/retrospection_examples.tex
\section{Retrospection Prompts and Examples}
\label{app:retrospection_examples}

We present two retrospections: one following a failed attempt and one
following a successful attempt.

Each example includes the complete system and user messages and the full
saved continuation, including the model's thinking before
\texttt{</think>}. Truncation markers in the user messages are part of
the context supplied to the model, not additional omissions for presentation.
The chat template supplies the opening \texttt{<think>} before generation.
The saved continuations retain nonempty thinking after the preprocessing
described in \cref{app:experimental_setup}; only the end-of-turn
suffix appended by the trainer is omitted below.
Verdict feedback is optional. Both saved examples include it; we display
this part of each user message in a separate box labeled ``Verdict
(optional)'', preserving the original prompt text.

\subsection{Failed attempt: database-session parameter collision}
\label{app:retrospection_failed}

The agent repairs a naming collision between an internal database-session
parameter and a model field. Renaming the parameter in \texttt{\_save}
does not address the same collision in \texttt{\_get\_or\_create},
and the latter path fails its test. The retrospection identifies the
incomplete coverage and recommends checking related entry points when
internal parameters can collide with user-supplied fields.

\begin{retrospectionrecord}{Failed attempt --- full system prompt}{retrospection-system}
\retrospectioninput{overleaf_rr_public/assets/retrospection/failed_system.txt}
\end{retrospectionrecord}
\begin{retrospectionrecord}{Failed attempt --- user prompt}{retrospection-user}
\retrospectioninput[lastline=247]{overleaf_rr_public/assets/retrospection/failed_user.txt}
\end{retrospectionrecord}
\begin{retrospectionrecord}{Failed attempt --- Verdict (optional)}{retrospection-user}
\retrospectioninput[firstline=248,lastline=315]{overleaf_rr_public/assets/retrospection/failed_user.txt}
\end{retrospectionrecord}
\begin{retrospectionrecord}{Failed attempt --- user prompt (continued)}{retrospection-user}
\retrospectioninput[firstline=316]{overleaf_rr_public/assets/retrospection/failed_user.txt}
\end{retrospectionrecord}
\begin{retrospectionrecord}{Failed attempt --- full retrospection, including thinking}{retrospection-target}
\retrospectioninput{overleaf_rr_public/assets/retrospection/failed_retrospection.txt}
\end{retrospectionrecord}

\subsection{Successful attempt: context-manager cleanup}
\label{app:retrospection_passed}

The agent fixes a context manager whose interaction state persists when a
test raises an exception. The usual verification step clears this state,
but exceptional exits bypass verification. The retrospection explains why
cleanup must also occur on that path and identifies failures in subsequent
tests as a signal of leaked state.

\begin{retrospectionrecord}{Passed attempt --- full system prompt}{retrospection-system}
\retrospectioninput{overleaf_rr_public/assets/retrospection/passed_system.txt}
\end{retrospectionrecord}
\begin{retrospectionrecord}{Passed attempt --- user prompt}{retrospection-user}
\retrospectioninput[lastline=679]{overleaf_rr_public/assets/retrospection/passed_user.txt}
\end{retrospectionrecord}
\begin{retrospectionrecord}{Passed attempt --- Verdict (optional)}{retrospection-user}
\retrospectioninput[firstline=680,lastline=749]{overleaf_rr_public/assets/retrospection/passed_user.txt}
\end{retrospectionrecord}
\begin{retrospectionrecord}{Passed attempt --- user prompt (continued)}{retrospection-user}
\retrospectioninput[firstline=750]{overleaf_rr_public/assets/retrospection/passed_user.txt}
\end{retrospectionrecord}
\begin{retrospectionrecord}{Passed attempt --- full retrospection, including thinking}{retrospection-target}
\retrospectioninput{overleaf_rr_public/assets/retrospection/passed_retrospection.txt}
\end{retrospectionrecord}

%% file: sections/appendix/generalization_efficiency.tex
\section{Generalization and Training Efficiency Experiments}
\label{app:generalization_efficiency}

\subsection{Experimental Protocol}

We compare \methodabbr{} and GRPO initialized from Qwen3.5-4B and
trained on SWE-rebench-767. Both methods use four NVIDIA B200 GPUs
for training and four for inference. Each \methodabbr{} update uses
64 solution attempts, with up to four retrospections per attempt.
GRPO instead uses 32 accepted groups of eight attempts, rejecting
groups with identical rewards because they provide no relative
advantage signal.

We measure training cost using elapsed time and the number of
completed solution attempts, including attempts rejected before
optimization. Retrospection generations are excluded from the attempt
count, but their cost is included in elapsed time. Evaluation time
and downtime between training sessions are excluded. At 40 updates,
\methodabbr{} requires \EfficiencyRRHours{} hours and
\EfficiencyRRSamples{} attempts, compared with
\EfficiencyGRPOHours{} hours and \EfficiencyGRPOSamples{} for GRPO.
The sample counts are taken from the latest completed-attempt reports
available at each checkpoint.

\subsection{Training Curves}

\Cref{fig:training-efficiency} reports training reward before
group rejection, rather than held-out task success. In particular,
GRPO's length penalty affects this reward. For both methods, we
smooth the curves over the preceding hour, five optimizer updates,
or 1,024 solution attempts, depending on the horizontal axis.
The time average weights observations by duration; the update and
sample averages weight them by their numbers of attempts.
Because generation is asynchronous, batches are assigned to update
intervals by completion-report time.

Each panel ends at the smaller of the two methods'
resource budgets. \Cref{fig:extended-training-checkpoints}
extends GRPO to 100 updates while \methodabbr{} ends at 40, with
the same one-hour smoothing applied throughout.

\subsection{Held-Out Performance}

We evaluate saved checkpoints on SWE-bench Verified separately from
training. \Cref{fig:extended-training-checkpoints} reports their
solve rates alongside the training curves. \methodabbr{} reaches 49.0\%
after ten updates and 1.29 hours, whereas GRPO reaches 48.0\% after
40 updates and 8.30 hours.

\paragraph{Cross-benchmark comparison.}
\Cref{fig:checkpoint-benchmark-bars} compares \methodabbr{} at
update 20 and GRPO at update 40. These checkpoints
solve 246/500 and 240/500 Verified tasks, respectively. Evaluating the
same checkpoints on SWE-bench Pro yields 212/731 and 185/731 solved
tasks (29.0\% and 25.3\%).

\paragraph{Training dynamics.}
\methodabbr{}'s smoothed training reward peaks at
\EfficiencyRRPeakReward{} near update \EfficiencyRRPeakUpdate{}
and declines to \EfficiencyRRFinalReward{} by update 40.
GRPO continues to improve its training reward with longer training,
but its Verified solve rate falls from 48.0\% at update 40 to
46.0\% at update 90. During \methodabbr{} training, SFT loss and
retrospection entropy also decrease.

%% file: sections/appendix/beyond_frontier_learning.tex
\section{Single-Task Training Experiments}
\label{app:beyond_frontier_learning}

\subsection{Single-Task Training and Measurement}
\label{app:single_task_protocol}

Each experiment starts from Qwen3.5-4B and repeatedly trains on a single
task. SymPy~20916, Django~15554, and Pylint~6386 are drawn from SWE-bench
Verified; SymbiFlow~17 is drawn from SWE-rebench. Each update uses
64 solution attempts and up to four self-generated retrospections per
attempt. We apply supervised fine-tuning only to retrospection tokens,
with a learning rate of $10^{-6}$. Solution attempts use temperature 1.0,
top-$p=1.0$, a 100-turn limit, a 131,072-token context, and at most
8,192 generated tokens per turn.
The initial training batches contain 0/64 successes for SymPy, SymbiFlow,
and Django, and 2/64 for Pylint.

The learning curves report the fraction of successful solution attempts
during training, before reward shaping or training-batch filtering.
Attempts excluded from training and execution failures assigned zero
reward remain in this measurement.
We smooth the curves by pooling the smallest set of consecutive, most
recent reporting intervals containing at least 1,024 attempts, or all
available attempts before that threshold.
Each rate is the total number of successes divided by the total
number of attempts in the window. Retaining whole reporting intervals
gives full windows of
\HardWindowMin{}--\HardWindowMax{} attempts.
All observed attempts contribute to smoothing; when multiple observations
share an update number, the latest observation is plotted.

\subsection{Additional Single-Task Learning Results}
\label{app:additional_task_learning}

\begin{figure}[htbp]
    \centering
    \begin{subfigure}[t]{0.49\textwidth}
        \includegraphics[width=\linewidth]{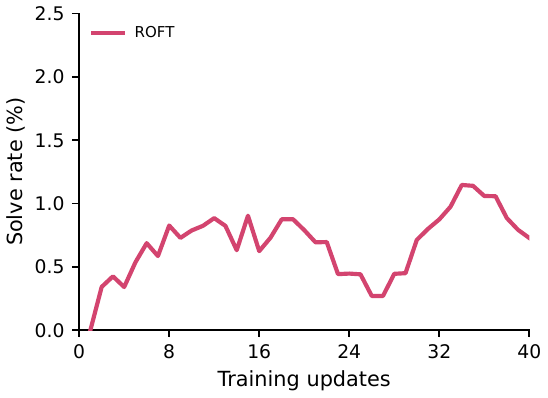}
        \caption{Django 15554: 40 updates.}
        \label{fig:additional-django}
    \end{subfigure}\hfill%
    \begin{subfigure}[t]{0.49\textwidth}
        \includegraphics[width=\linewidth]{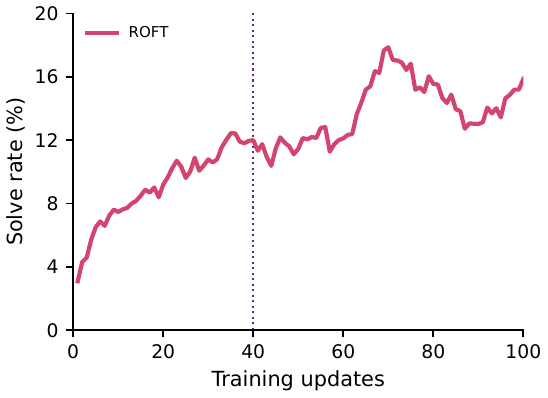}
        \caption{Pylint 6386: 100 total updates.}
        \label{fig:zero-advantage-pylint}
    \end{subfigure}
    \caption{Online solve rates for Django through 40 updates and Pylint
    through 100 updates, using the smoothing procedure in
    \cref{app:single_task_protocol}.
    The dotted line at update 40 marks the checkpoint from which Pylint
    training resumes. Vertical scales differ.}
    \label{fig:additional-task-learning}
\end{figure}

\Cref{fig:additional-task-learning} shows the Django experiment and
the longer Pylint trajectory. Django's smoothed solve rate peaks at
\HardDjangoPeak{} and ends at \HardDjangoFinal{} after 40 updates.
For Pylint, we resume the 40-update checkpoint for 60 further updates,
giving 100 updates in total. Its smoothed solve rate peaks at
\HardPylintPeak{} and ends at \HardPylintFinal{}.
The first 40 updates are shown in \cref{fig:low-pass-rate-pylint}.

\begin{figure}[htbp]
    \centering
    \includegraphics[width=0.52\textwidth]{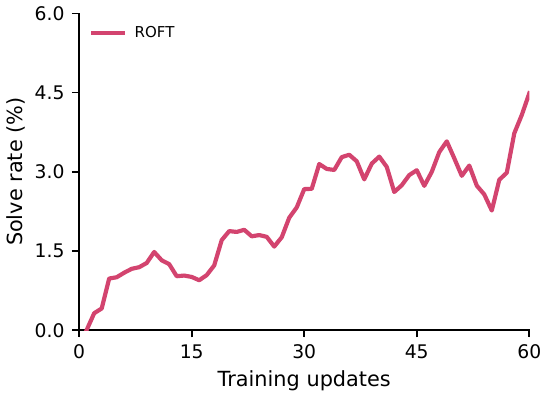}
    \caption{Online solve rate for SymbiFlow~17 through 60 updates,
    extending the trajectory in \cref{fig:zero-advantage-symbiflow}
    with the same smoothing procedure.}
    \label{fig:additional-symbiflow-60}
\end{figure}

\Cref{fig:additional-symbiflow-60} extends the main-text SymbiFlow
trajectory to 60 updates of the same run. Its smoothed solve rate reaches
\HardSymbiflowExtendedFinal{} at update 60, the highest value within
this interval.

\subsection{Test-Set Performance after Single-Task Training}
\label{app:single_task_test_performance}

\begin{figure}[htbp]
    \centering
    \includegraphics[width=0.88\textwidth]{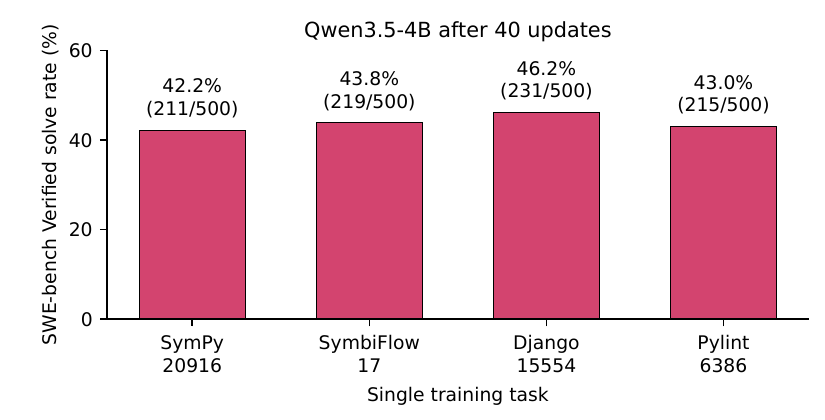}
    \caption{SWE-bench Verified performance after 40 updates of
    single-task \methodabbr{} from Qwen3.5-4B. Each bar reports an
    evaluation over all 500 tasks; labels give the solve rate and
    solved/total count. The single training instance is not excluded
    for the three runs trained on Verified tasks. Execution
    failures count as unresolved.}
    \label{fig:single-task-test-performance}
\end{figure}

\Cref{fig:single-task-test-performance} reports SWE-bench Verified
solve rates for the four single-task runs after 40 optimizer updates.
Each frozen checkpoint is evaluated on all 500 tasks with one
attempt per task, using temperature 1.0, top-$p=1.0$,
a 100-turn limit, a 131,072-token context, and at most 8,192
generated tokens per turn. Retrospection and context compaction
are disabled during evaluation.

We retain all 500 tasks in each denominator, counting execution
failures as unresolved.
The SymPy, Django, and Pylint training instances remain in the
evaluation set. SymbiFlow's training instance is from SWE-rebench.

%% file: sections/appendix/behavioral_induction.tex
\section{Behavioral Analyses}
\label{app:behavioral_induction}

\subsection{Likelihood Changes by Turn Correctness}
\label{app:credit_assignment}

\begin{figure}[!htb]
    \centering
    \includegraphics[width=\linewidth]{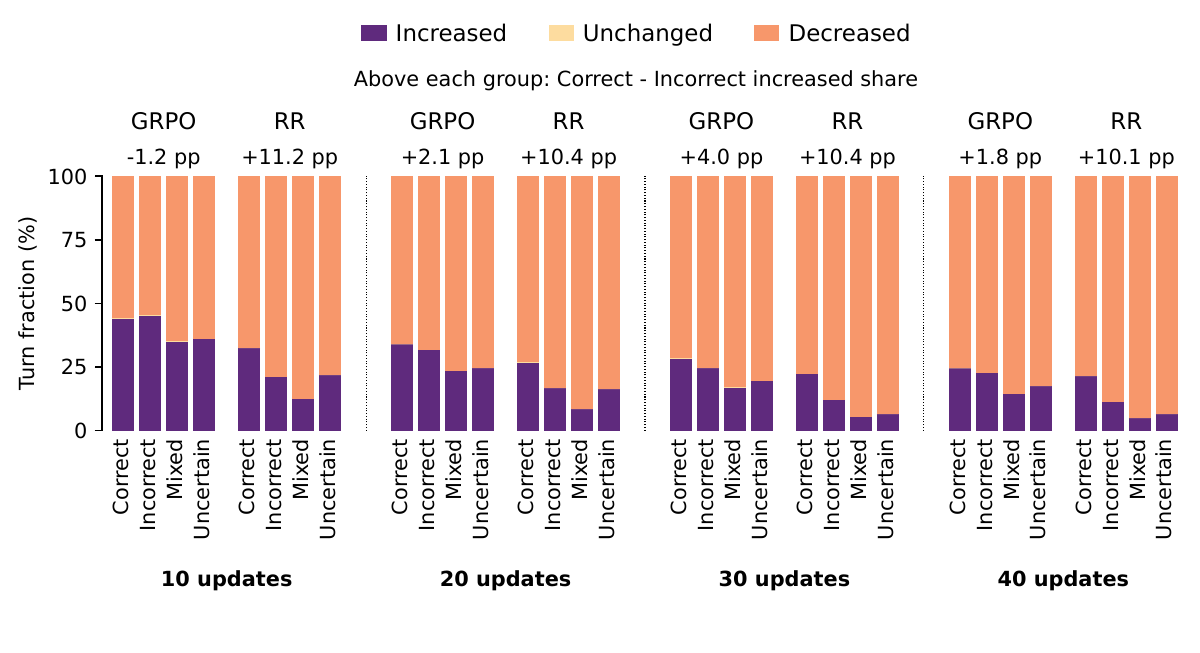}
    \caption{\textbf{Likelihood-change directions across training.}
    Each update group compares GRPO with retrospection-only training
    (RR) on the same recorded turns. Bars show trajectory-balanced
    fractions for each correctness label: increased likelihood
    (purple), effectively unchanged likelihood (light yellow), and
    decreased likelihood (orange), relative to the initial model.
    Signed annotations give the correct-minus-incorrect increased
    fraction in percentage points.
    \Cref{fig:credit-assignment} displays the correct and
    incorrect increased fractions at update 10.}
    \label{fig:credit-assignment-full}
\end{figure}

\Cref{fig:credit-assignment-full} compares how retrospection-only
training (RR) and GRPO change the likelihood of recorded assistant
turns, grouped by correctness. We evaluate both methods after
10, 20, 30, and 40 optimizer updates, using the same turns and
the initial Qwen3.5-4B model as the reference.

\paragraph{Trajectory labeling.}
We use 256 trajectories generated by the initial model on
SWE-rebench-767 and retained in the first accepted batch of an
earlier GRPO run. Each turn comprises the assistant's reasoning
and actions. GPT-6 Astra labels every turn using the complete task,
trajectory, tool observations, submitted patch, and final verdict.
The labeling rubric distinguishes \emph{correct} turns, whose
substantive reasoning and actions are supported by the evidence and
valid for solving or checking the task;
\emph{incorrect} turns, for which the evidence establishes a
substantive error; \emph{mixed} turns, containing both correct and
incorrect elements; and \emph{uncertain} turns, for which the record
is insufficient to judge. The judge has no access to checkpoint scores
or external tools. We validate turn coverage and evidence references, then hold
the labels fixed across methods and checkpoints.
Two trajectories exceed the judge's context limit and are excluded,
leaving 254 trajectories with 14,200 labeled turns:
\StepCorrectTurns{} correct, \StepIncorrectTurns{} incorrect,
\StepMixedTurns{} mixed, and \StepUncertainTurns{} uncertain.

\paragraph{Scoring likelihood changes.}
For each checkpoint, we teacher-force the original tokens using the
same recorded histories. For a turn $s$ containing $n_s$ assistant
tokens $x_{s,1:n_s}$ after history $h_s$, we compute the change in
mean token log-likelihood:
\begin{equation}
    \Delta_s^{(u)}
    = \frac{1}{n_s}\sum_{j=1}^{n_s}
      \left[
      \log \pi_u(x_{s,j}\mid h_s,x_{s,<j})
      - \log \pi_0(x_{s,j}\mid h_s,x_{s,<j})
      \right],
    \label{eq:step_likelihood_delta}
\end{equation}
where $\pi_0$ is the initial model and $\pi_u$ is the checkpoint
after $u$ updates. The history contains only the task and preceding
messages and observations; neither future evidence nor correctness
labels are supplied during scoring. We evaluate assistant-token
probabilities with dropout disabled and temperature one, using
BF16 logits and FP32 log-softmax reductions.
A turn's likelihood is classified as \emph{increased} if
$\Delta_s^{(u)}>10^{-6}$,
\emph{decreased} if $\Delta_s^{(u)}<-10^{-6}$, and
\emph{effectively unchanged} otherwise, in nats per assistant token.

\paragraph{Aggregation.}
For each method, checkpoint, and correctness label, we compute the
fraction of turns in each likelihood-change category within each
trajectory. We then average these fractions equally over trajectories
containing that label, so that longer trajectories do not receive
greater weight. Each stacked bar in
\cref{fig:credit-assignment-full} shows the resulting increased,
unchanged, and decreased fractions, which sum to 100\%.
The annotation above each method gives the increased fraction for
correct turns minus that for incorrect turns, in percentage points.
\Cref{fig:credit-assignment} uses the same calculation but shows
only the correct and incorrect increased fractions at update 10,
without renormalization.

\paragraph{Results.}
Under RR, correct turns increase in likelihood more often than
incorrect turns at all four checkpoints. The correct-minus-incorrect
gap is larger for RR than for GRPO throughout:
\StepRRMainGap{} versus $\StepGRPOMainGap{}$ percentage points at
update 10, and \StepRRFinalGap{} versus \StepGRPOFinalGap{} at
update 40. At update 40, the correct and incorrect increased
fractions are
\StepRRCorrectIncreased{} and \StepRRIncorrectIncreased{},
respectively, for RR, compared with \StepGRPOCorrectIncreased{} and
\StepGRPOIncorrectIncreased{} for GRPO.
At this checkpoint, the majority of turns in each of the four
correctness categories decrease in likelihood under both methods.

\input{sections/appendix/faster_rollouts}

\subsection{Fine-Grained Trajectory and Retrospection-Token Analyses}
\label{app:reflection_token_analysis}

\paragraph{Training batch and one-update checkpoints.}
We use 64 fixed SWE-rebench-767 problems and one original Qwen3.5-4B
trajectory per problem. The trajectories contain 32 successful and
32 unsuccessful attempts. Four retrospections were requested per
trajectory; 255 were accepted and one generation was rejected.
Retrospections were sampled at temperature 0.9 with a
24,576-token input budget. The frozen batch has
214,072 supervised target tokens.
The Uniform checkpoint performs exactly one globally token-normalized
retrospection-only update from the base model, using Adam with learning
rate $10^{-6}$ and gradient clipping at 1. The task and recorded attempt
are conditioning context; only the retrospection targets incur
prediction loss. The weighting variants described below retain the same
base initialization, target batch, optimizer settings, and update count.

\paragraph{Fixed-prefix distribution comparison.}
For the Uniform checkpoint, we score the 64 original trajectories,
comprising 3,604 assistant spans and 1,085,552 assistant-token positions,
under both the base and updated models. Each trajectory is included
once, independently of its number of accepted retrospections.
At assistant position $j$, let $h_j$ be the original causal prefix and
$x_j$ the recorded token. We compute
\begin{equation}
    K_j = \sum_{v\in\mathcal{V}}\pi_0(v\mid h_j)
          \log\frac{\pi_0(v\mid h_j)}{\pi_1(v\mid h_j)},
    \qquad
    d_j = \log\pi_1(x_j\mid h_j)-\log\pi_0(x_j\mid h_j),
    \label{eq:reflection_token_metrics}
\end{equation}
where $\pi_0$ and $\pi_1$ denote the base and one-update Uniform
policies. The full-vocabulary comparison uses saved BF16 logits and
FP32 normalizers; all reported values retain their raw signs.
The prefixes are identical across checkpoints, with no tool
reexecution or newly sampled actions during scoring.

\begin{figure}[htbp]
    \centering
    \includegraphics[width=\linewidth]{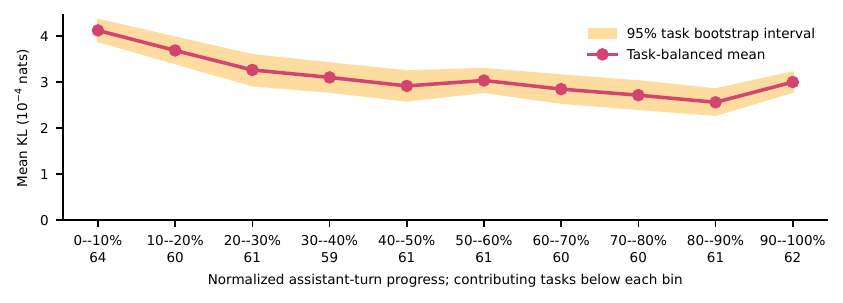}
    \caption{\textbf{Distributional change across rollout progress.}
    Task-balanced forward KL between the base and one-update Uniform
    policies, grouped by normalized assistant-turn progress.
    The band shows the 95\% task-bootstrap interval.
    Counts below the bins give the number of contributing tasks.}
    \label{fig:reflection-progress}
\end{figure}

\paragraph{Mechanical grouping and aggregation.}
We group positions into reasoning, text, tool-call arguments,
format delimiters, tool-call syntax, and unknown roles using the
serialized rollout text. These labels are distinct from the semantic
retrospection labels used for loss weighting.
Tool groups preserve the serialized tool name and include reasoning
preceding the call; argument tokens in multi-tool calls are assigned
to their own tool. Unsupported parameter markup remains unknown.
For each category, we first average $K_j$ over its tokens within
each contributing task, then average the task means equally.
Intervals use 10,000 task-bootstrap resamples.
The tool panel in \cref{fig:reflection-token-analysis} displays
categories with at least ten contributing tasks.
For the progress analysis, we use ten equal-width bins of
normalized assistant-turn progress, applying the same within-task
then across-task averaging. Counts for each bin appear in
\cref{fig:reflection-progress}.

\paragraph{Distributional results.}
The global task-balanced mean KL is $3.27893\times10^{-4}$ nats,
with 95\% interval $[3.07268,3.50592]\times10^{-4}$.
The token-weighted mean is $3.11594\times10^{-4}$ nats,
the median is $1.74063\times10^{-6}$, and the maximum is 0.778787.
Task-balanced role means, in units of $10^{-4}$ nats, are
5.71843 for reasoning (64 tasks), 3.12641 for text (64),
1.65515 for arguments (62), 0.43545 for delimiters (64),
and 0.24240 for call syntax (62). The unknown-role category
contains 1,001 tokens from five tasks and has mean 8.62500 in
the same units.
Tool means are 4.30960 for \texttt{Glob} (56 tasks),
3.73144 for \texttt{Read} (62), 3.19321 for \texttt{Bash} (58),
2.75738 for \texttt{Write} (14), 2.27982 for \texttt{Edit} (58),
and 3.72441 for spans without a tool call (52).
The command-category means are 3.82981 for read/search commands
and 2.53292 for edits.
The first progress decile has mean 4.12157, compared with 2.99964
in the last decile, again in units of $10^{-4}$ nats.
The minimum estimated KL is $-3.97947\times10^{-6}$ nats:
226,337 positions have negative estimates, and 608,045 positions
(56.0\%) have absolute KL at or below its magnitude. No negative
values are clamped in the summaries or token windows.

\paragraph{Retrospection-token interventions.}
Uniform assigns raw weight 1 to every supervised target token.
Thinking-down and thinking-up assign raw weight 0.5 or 2,
respectively, to mechanically identified thinking tokens and 1
elsewhere. Evidence-up, correction-up, and lesson-up each assign
raw weight 2 to tokens in the selected semantic category and 1
elsewhere. Evidence denotes observations, test results, errors,
or concrete behavior supporting or contradicting a decision;
correction denotes a task-specific replacement assumption,
action, or fix; lesson denotes a reusable decision rule or trigger.
GPT-6 Astra annotations cover all 255 accepted retrospections.
The fixed character-span labels are mapped onto the target tokens.
The selected categories contain
71,226 evidence tokens, 25,591 correction tokens, and 21,347 lesson
tokens. Other, mixed, uncertain, boundary-neutral, and control
tokens retain raw weight 1 in these category-based variants.

Attention-up uses the original base model's attention to the user-prompt
content, which includes the recorded trajectory. For each nonstructural
retrospection target token, we measure the total attention mass assigned
to that content at the causal position immediately preceding the token,
then average over 16 query heads and eight full-attention layers.
If this mean mass is $m_i$, the token receives raw weight $a_i=1+m_i$;
structural control tokens retain raw weight 1. Unlike the category-based
doubling interventions, this is a continuous weighting rule.
The scores are fixed before training, measured before the attention output
gate, and exclude the model's 24 linear-attention layers. 

For $N$ supervised tokens with raw weights $a_i$, each intervention uses
\begin{equation}
    \widetilde{a}_i =
    \frac{a_i}{N^{-1}\sum_{k=1}^{N}a_k},
    \qquad
    \mathcal{L}_{\mathrm{weighted}} =
    -\frac{1}{N}\sum_{i=1}^{N}
       \widetilde{a}_i\log\pi_\theta(y_i\mid h_i).
    \label{eq:reflection_weighted_loss}
\end{equation}
Here $h_i$ contains the task, recorded attempt, supplied feedback,
and preceding retrospection tokens for target position $i$.
Normalization is global across the target batch, not
per retrospection or per semantic category. The targets and
conditioning contexts are unchanged.

\paragraph{Fresh solving evaluations.}
We compare Uniform with six reweighted variants, each a fixed one-update
checkpoint evaluated on the same 64 problems with ten repetitions.
This comparison includes 640 graded attempts per model and 4,480 in total.
Attention-up comes from a follow-up experiment that reuses the original
Uniform evaluation, with identical task--seed assignments and
inference settings; the control is not rerun or counted twice.
Randomization is paired by task and repetition across models. Decoding uses
temperature 1, at most 8,192 generated tokens per turn, 100 turns,
and 131,072 context tokens. No new training or retrospection
generation occurs during these evaluations, and saved retrospections
are not included in the solver context.

\paragraph{Performance estimator and standard errors.}
The solve rate is the mean binary outcome over all 640 attempts,
equivalently the average of the 64 task-specific ten-attempt means.
Contrasts against Uniform pair task and repetition IDs. The fixed-panel bootstrap
resamples ten paired repetitions independently within each task,
then averages over all 64 tasks, using 10,000 bootstrap resamples.
The fixed-panel error bars in \cref{fig:reflection-token-analysis}
quantify uncertainty in the mean paired difference on these fixed tasks;
they show the observed mean plus or minus one bootstrap standard error.
We estimate this standard error as the sample standard deviation of the
10,000 resampled mean differences, using denominator 9,999.
\Cref{tab:reflection_weighting_results} reports scores for Uniform and
the six reweighted variants, with these same standard errors for
changes relative to Uniform.
Attention-up solves 350/640 problems (54.69\%), compared with
324/640 (50.63\%) for Uniform, a $+4.06$ percentage-point change.
Correction-up and evidence-up yield $+3.75$ and $+3.59$ points.

\begin{table}[htbp]
    \centering
    \small
    \input{overleaf_rr_public/reflection64/paper-table}
    \caption{\textbf{Repeated-evaluation results on the fixed 64 training problems.}
    Rates average ten attempts per task. Changes relative to Uniform and
    their standard errors (SE) are in percentage points. SEs resample paired
    repetitions independently within each of the 64 fixed tasks.}
    \label{tab:reflection_weighting_results}
\end{table}

\subsection{Token-Level Examples}
\label{app:reflection_token_examples}

\begin{figure}[htbp]
    \centering
    \includegraphics[width=\linewidth]{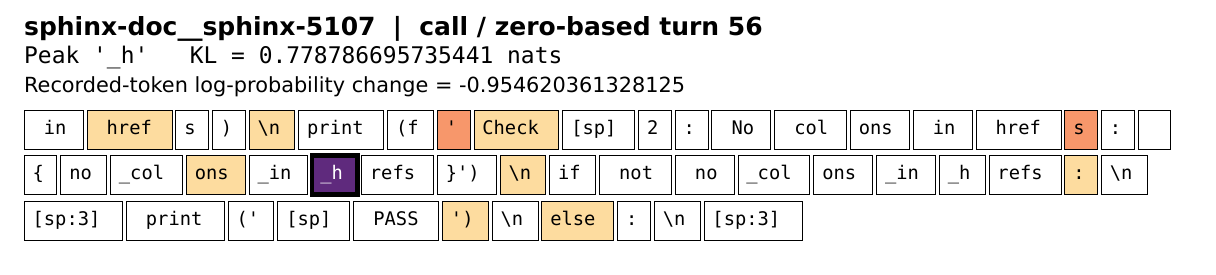}
    \smallskip
    \includegraphics[width=\linewidth]{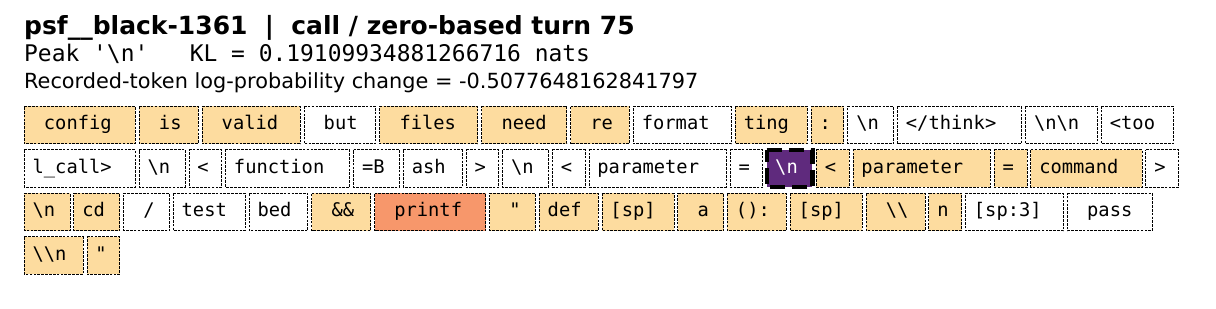}
    \smallskip
    \includegraphics[width=\linewidth]{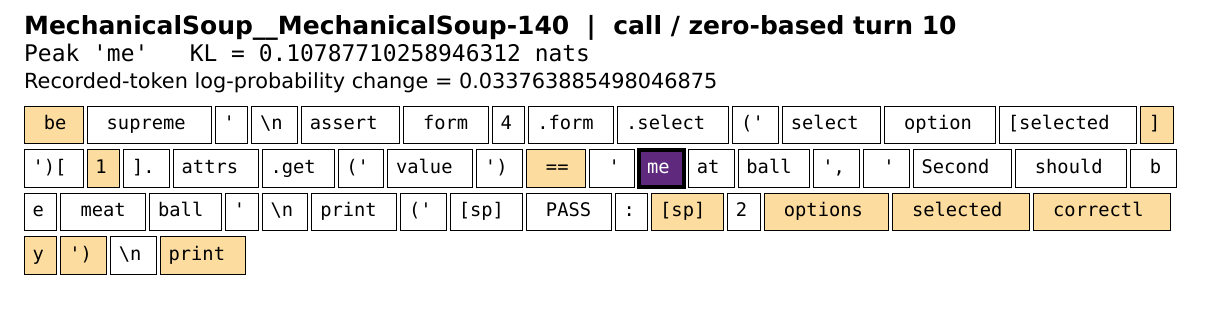}
    \includegraphics[width=\linewidth]{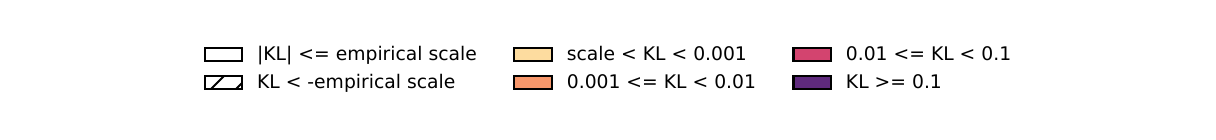}
    \caption{\textbf{Token-level forward-KL measurements.}
    Original 49-token rollout windows for \texttt{sphinx-doc/sphinx-5107}
    (top), \texttt{psf/black-1361} (middle), and
    \texttt{MechanicalSoup/MechanicalSoup-140} (bottom), colored by
    forward KL in nats after one Uniform update. Thick borders mark the
    selected peaks; dashed borders indicate non-exact parsing metadata.
    The Black peak has unknown tool and role labels because its parameter
    markup is malformed. The Sphinx and MechanicalSoup peaks are parsed
    \texttt{Bash} arguments. White cells have absolute KL at or below the
    magnitude of the minimum recorded KL, $3.98\times10^{-6}$ nats.}
    \label{fig:reflection-token-examples}
\end{figure}

\paragraph{Selection and measurements.}
We select each original trajectory's highest raw forward-KL
position, rank those peaks by descending KL, and break ties by
original dataset order. \Cref{fig:reflection-token-examples} shows
the three highest peaks, with 24 tokens on either side of each
selected position. These are original action-trajectory tokens, not
retrospection targets. We use the fixed-prefix measurements defined in
\cref{eq:reflection_token_metrics}.

\paragraph{Sphinx: a generated check for EPUB links.}
The \texttt{sphinx-doc/sphinx-5107} window comes from a \texttt{Bash}
command that writes a Python script to build EPUB documentation and
inspect whether internal link fragments match their target identifiers
after colon-to-hyphen conversion. The selected token is \texttt{\_h}
inside \texttt{no\_colons\_in\_hrefs}, in a diagnostic print statement,
at zero-based assistant turn 56, token offset 1195.
It has the largest recorded forward KL in the study:
$K_j=0.77878670$ nats. Its recorded-token log-probability change is
$d_j=-0.95462036$ nats, while full-distribution entropy increases
by 1.11707091 nats. The saved script defines the variable to be
true when no link fragments contain colons, but prints \texttt{PASS}
when that condition is false. The selected peak occurs in the
preceding print statement.

\paragraph{Black: malformed tool markup.}
For \texttt{psf/black-1361}, the selected newline is at zero-based
assistant turn 75, token offset 55: $K_j=0.19109935$ nats and
$d_j=-0.50776482$. Its parameter markup is unparsed, and its tool
and role labels are unknown.

\paragraph{MechanicalSoup: a value in a test assertion.}
For \nolinkurl{MechanicalSoup/MechanicalSoup-140}, the selected
\texttt{me} token in \texttt{meatball} is at turn 10, offset 781:
$K_j=0.10787710$ nats and $d_j=+0.03376389$.
It is a \texttt{Bash} argument within an assertion on the second
selected option's value.

%% file: sections/appendix/faster_rollouts.tex
\subsection{Rollout Lengths with Efficiency-Focused Retrospections}
\label{app:faster_rollouts}

We measure coding-rollout lengths under two retrospection instructions:
our default prompt and a variant requesting more direct solutions.
The task-solving prompt and training objective remain unchanged.

\paragraph{Experimental procedure.}
We compare two \methodabbr{} runs initialized from the same Qwen3.5-4B
checkpoint and trained on the same 767-task dataset.
Each run performs 40 optimizer updates, using 64 task attempts and
four independently sampled retrospections per attempt in each batch.
Both successful and unsuccessful attempts contribute training targets.
The loss is averaged over retrospection tokens only, with no
supervision on task-solving tokens and no length penalty.
Both runs use a constant learning rate of $10^{-6}$.
Task solving uses temperature 1.0, a limit of 100 assistant turns,
and at most 8,192 generated tokens per turn.
Retrospections use temperature 0.9, an input budget of 24,576 tokens,
and an output limit of 8,192 tokens.

Our default prompt asks the model to identify a consequential
assumption or decision, explain the relevant evidence, and state a
lesson for future attempts.
The faster variant instead selects its instruction according to
the final test verdict.
After a successful attempt, it asks the model to identify avoidable
work, such as repeated searches or unnecessary edits, and propose a
more direct approach while preserving the checks needed for correctness.
After an unsuccessful attempt, it asks for a wrong assumption or
decision supported by the evidence and a concrete correction.
In both conditions, the retrospection receives the task, recorded
trajectory, submitted patch, and test verdict.
The complete system instructions are reproduced below.

\paragraph{Rollout-length measurement.}
\Cref{fig:faster-rollouts} compares training rollouts around the
tenth update, pooling batches 8--12 for 320 attempts per condition.
We measure generated assistant tokens, including reasoning and actions,
and the number of assistant turns in each original coding trajectory.
Retrospection generations are excluded, and each attempt contributes
once. We report means over all attempts and separately over solved
and unsolved attempts, as determined by the final test verdict.
Attempts that reach a generation budget remain included.
Means weight individual attempts equally, rather than weighting
batches or outcome groups equally.

\begin{table}[htbp]
    \centering
    \small
    \caption{Mean coding-rollout lengths around update 10, pooled over
    batches 8--12. Percentage changes compare the faster-retrospection
    variant with the baseline. Each attempt receives equal weight.}
    \label{tab:faster-rollouts}
    \input{overleaf_rr_public/assets/faster_prompt/early_window_table}
\end{table}

\paragraph{Results.}
The faster variant produces shorter rollouts in all three groups
(\cref{tab:faster-rollouts}).
Across all attempts, it uses \FasterAllTokenReduction{} fewer tokens
and \FasterAllTurnReduction{} fewer turns than the baseline.
The reductions also hold among solved attempts:
\FasterSolvedTokenReduction{} fewer tokens and
\FasterSolvedTurnReduction{} fewer turns.
Among unsolved attempts, token and turn counts decrease by
\FasterUnsolvedTokenReduction{} and \FasterUnsolvedTurnReduction{},
respectively.
The solve rates in this window are \FasterBaselinePassRate{} for the
baseline and \FasterTreatmentPassRate{} for the faster variant.

\begin{retrospectionrecord}{Default prompt --- both verdicts}{retrospection-system}
\retrospectioninput{overleaf_rr_public/assets/faster_prompt/baseline-system.txt}
\end{retrospectionrecord}

\begin{retrospectionrecord}{Faster variant --- passing verdict}{retrospection-system}
\retrospectioninput{overleaf_rr_public/assets/faster_prompt/faster-passed-system.txt}
\end{retrospectionrecord}

\begin{retrospectionrecord}{Faster variant --- failing verdict}{retrospection-system}
\retrospectioninput{overleaf_rr_public/assets/faster_prompt/faster-failed-system.txt}
\end{retrospectionrecord}

%% file: overleaf_rr_public/assets/faster_prompt/early_window_table.tex
\begin{tabular}{@{}lrrrrrr@{}}
\toprule
& \multicolumn{3}{c}{Mean assistant tokens} & \multicolumn{3}{c}{Mean assistant turns} \\
\cmidrule(lr){2-4}\cmidrule(l){5-7}
Rollouts & Baseline & Faster & Change & Baseline & Faster & Change \\
\midrule
All & 12,674 & 11,186 & -11.7\% & 47.05 & 40.88 & -13.1\% \\
Solved & 13,064 & 11,454 & -12.3\% & 48.03 & 41.83 & -12.9\% \\
Unsolved & 12,050 & 10,808 & -10.3\% & 45.48 & 39.56 & -13.0\% \\
\bottomrule
\end{tabular}

%% file: overleaf_rr_public/reflection64/paper-table.tex
\begin{tabular}{lrrrr}\toprule
Model & Solved & Rate & $\Delta$ vs.\ Uniform & SE (pp) \\
\midrule
Uniform & 324/640 & 50.63\% & -- & -- \\
Thinking-down & 318/640 & 49.69\% & -0.94 & 2.28 \\
Thinking-up & 330/640 & 51.56\% & +0.94 & 2.23 \\
Evidence-up & 347/640 & 54.22\% & +3.59 & 2.26 \\
Correction-up & 348/640 & 54.38\% & +3.75 & 2.12 \\
Lesson-up & 344/640 & 53.75\% & +3.13 & 2.15 \\
Attention-up & 350/640 & 54.69\% & +4.06 & 2.17 \\
\bottomrule\end{tabular}

%% file: sections/appendix/ablations.tex
\section{Ablations}
\label{app:ablations}

The Qwen3.5-4B ablations initialize from the same base checkpoint and
train on SWE-rebench-767. Each optimizer update uses 256 nominal
retrospection targets and minimizes token-normalized cross-entropy
with a constant learning rate of $10^{-6}$. The conditioning context
is masked, and only retrospection tokens incur loss.

\subsection{Retrospection with and without a Verdict}
\label{app:verdict_ablation}

We compare verdict-conditioned retrospection with a variant that omits
the final grader feedback. Each optimizer update uses 64 source coding
attempts and four independently sampled retrospections per attempt.

The verdict-conditioned run is the baseline used in the main
generalization comparison. 
Its retrospection context includes the
task, trajectory evidence, patch, and final verdict.
The no-verdict run supplies only the final patch and a chronological
trajectory summary. It omits the separate task description, final outcome
label, reward values, termination metadata, grader-selected test
targets, and final grader logs. Pre-grading reasoning, actions, and
tool observations, including tests executed by the solving agent,
remain in the trajectory. The same fitted conditioning context is used for
generation and masked SFT. Source attempts are still graded for
metrics and validity handling; grades neither select the
retrospection instruction nor supply loss weights.

\begin{figure}[htbp]
    \centering
    \includegraphics[width=0.42\textwidth]{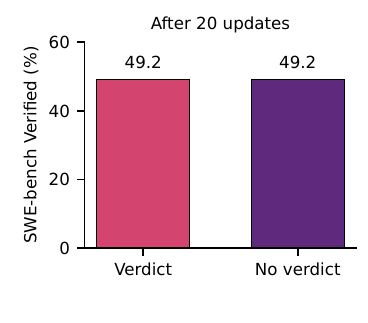}
    \caption{\textbf{Retrospection with and without a final verdict.}
    SWE-bench Verified solve rates after 20 optimizer updates from
    Qwen3.5-4B on SWE-rebench-767. Each bar reports the percentage of
    all 500 tasks resolved:
    \VerdictControlScore{} with a verdict and \VerdictRemovedScore{}
    without one (246 tasks in each case).}
    \label{fig:ablation-verdict}
\end{figure}

We evaluate both checkpoints after 20 successful optimizer updates,
with top-$p=1.0$ and at most 8,192 generated tokens per turn.
As shown in \cref{fig:ablation-verdict}, both checkpoints resolve
246/500 tasks, yielding \VerdictControlScore{} with the verdict and
\VerdictRemovedScore{} without it.

\subsection{Online, Off-Policy Reflections, and Offline RR}
\label{app:policy_ablation}

\Cref{fig:ablation-policy} compares three ways of constructing
retrospection supervision.

\paragraph{Online RR.}
The control is the verdict-conditioned run in
\cref{app:verdict_ablation}. The evolving learner generates both
coding attempts and retrospections.
Updated learner weights are used for subsequent solving and reflecting.
Generation is asynchronous; rollouts can begin under earlier
optimizer states.

\paragraph{Off-policy reflections.}
Coding attempts still come from the evolving learner and receive the
ordinary environment verdict. A separate, immutable copy of the
original base model generates all four retrospections for each attempt.
It receives that learner attempt's task, trajectory evidence, patch,
and verdict; it does not generate a replacement coding attempt.
Only the learner is optimized.
The eight-B200 allocation is split into four training GPUs, three
learner-serving GPUs, and one frozen-reflector GPU, rather than the
control's four training and four shared inference GPUs.

\paragraph{Offline RR.}
We first generate a fixed dataset without any learner updates.
For each of the 767 training problems, the frozen base model attempts
two independent solutions; after grading each attempt, that same base
model samples four retrospections.
After filtering invalid generations and excluding attempts with no
model response, the corpus contains 6,066 retrospections.
A separately initialized learner trains on this immutable
corpus in shuffled batches of 256 targets, reusing examples if needed.
No solving or retrospection generation is refreshed during training;
the corpus remains fixed throughout.

\paragraph{Checkpoints and evaluation.}
The online control and off-policy-reflection arm use the checkpoints
after 20 successful optimizer updates. 
The online result is 246/500 (\PolicyOnlineScore{}), as in the main comparison.
Off-policy reflections and offline RR solve 237/500
(\PolicyOffPolicyScore{}) and 240/500 (\PolicyOfflineScore{}).

\subsection{Number of Retrospections per Rollout}
\label{app:reflection_count}

We compare 2, 4, and 8 independently sampled retrospections per source
rollout. The respective
source batch sizes are 128, 64, and 32, yielding 256 nominal retrospection
targets per optimizer update. 
Empty, failed, or truncated retrospections are dropped
rather than duplicated to fill the batch, so the accepted target
count can be smaller than 256.

\Cref{fig:ablation-reflections} evaluates each run after ten
successful optimizer updates. Evaluations allow at most
8,192 generated tokens per turn.
The 2-, 4-, and 8-retrospection settings solve 239, 245, and 231 problems,
respectively.

The 4-retrospection control has the highest observed
score, and the three scores span \ReflectionScoreSpread{} percentage
points.

\subsection{Scaling to Qwen3.5-9B}
\label{app:nine_b}

We initialize both methods from Qwen3.5-9B and train on
SWE-rebench-767 with four training and four inference B200 GPUs.
\methodabbr{} uses 64 source attempts and four retrospections per
attempt; GRPO uses 32 accepted groups of eight attempts.
Both allow generation to lag the learner by at most three updates,
use at most 8,192 generated tokens per coding turn, and apply no
assistant-length penalty. RR samples retrospections at temperature 0.9
with a 24,576-token input budget.
GRPO trains for 40 updates and \methodabbr{} for 20 updates.
Training-cost accounting follows \cref{app:generalization_efficiency}.

At 20 updates, \methodabbr{} uses \NineBRRHours{} hours and
\NineBRRSamples{} completed solution attempts, compared with
\NineBGRPOHours{} hours and \NineBGRPOSamples{} for GRPO.
The corresponding checkpoints solve 294/500 and 279/500 SWE-bench
Verified tasks (\NineBRRScore{} and \NineBGRPOScore{};
\cref{fig:ablation-nine-b}).
